\pdfoutput=1
\documentclass[11pt]{article}

\usepackage[final]{acl}
\usepackage{times}
\usepackage{latexsym}

\usepackage[T1]{fontenc}
\usepackage[utf8]{inputenc}

\usepackage{microtype}

\usepackage{inconsolata}

\usepackage{graphicx, caption, subcaption}
\usepackage{enumitem}
\setlist[itemize]{leftmargin=14pt}

\usepackage[colorinlistoftodos,prependcaption,textsize=small]{todonotes}
\presetkeys{todonotes}{inline}{}

\usepackage{booktabs}
\usepackage{multirow}

\usepackage{xspace}
\usepackage{bbm}
\newcommand*{\LACOAT}{\texttt{LACOAT}}
\newcommand*{\mone}{\texttt{ConceptDiscoverer}}
\newcommand*{\mtwo}{\texttt{PredictionAttributor}}
\newcommand*{\mthree}{\texttt{ConceptMapper}}
\newcommand*{\mfour}{\texttt{PlausiFyer}}

\usepackage{amsfonts}
\usepackage{amsmath}
\usepackage{pifont}
\newcommand{\modelM}{\mathbb{M}}
\newcommand{\dataD}{\mathbb{D}}

\renewcommand*\cite[1]{\citep{#1}}

\title{Latent Concept-based Explanation of NLP Models
\\\hphantom{...}
\\\footnotesize{\textcolor{red}{WARNING: The appendix contains some examples, which may be disturbing to the reader.}}}
\author{
  Xuemin Yu$^{\diamondsuit}$,
  Fahim Dalvi$^{\clubsuit}$,
  Nadir Durrani$^{\clubsuit}$,
  Marzia Nouri$^{\heartsuit}$,
  Hassan Sajjad$^{\diamondsuit}$ \\
  $^{\diamondsuit}$Faculty of Computer Science, Dalhousie University, Canada \\
  $^{\clubsuit}$Qatar Computing Research Institute, Hamad Bin Khalifa University, Qatar \\
  $^{\heartsuit}$Independent Researcher \\
  \texttt{xuemin.yu@dal.ca}, \texttt{faimaduddin@hbku.edu.qa}, \\
  \texttt{ndurrani@hbku.edu.qa}, \texttt{nouri.marzia.1999@gmail.com}, \\
  \texttt{hsajjad@dal.ca}
}

\begin{document}
\maketitle

\begin{abstract}
Interpreting and understanding the predictions made by deep learning models poses a formidable challenge due to their inherently opaque nature. Many previous efforts to explain these predictions rely on input features, specifically, the words within NLP models. However, such explanations are often less informative due to the discrete nature of the words and their lack of contextual verbosity. To address this limitation, we introduce Latent Concept Attribution (\LACOAT{}), which generates explanations for predictions based on latent concepts. Our intuition is that a word can exhibit multiple facets depending on the context in which it is used. Therefore, given a word in context, the latent space derived from our training process reflects a specific facet of that word. \LACOAT{} functions by mapping the representations of salient input words into the training latent space, enabling it to provide latent  context-based explanations of the prediction.
\footnote{The codebase 
is available at \url{https://github.com/xuemin-yu/eraser_movie_latentConcept}.} 
%for reproducibility and 
\end{abstract}

\section{Introduction}
The opacity of deep neural network (DNN) models is a major challenge in ensuring a safe and trustworthy AI system. Extensive research works have attempted to interpret and explain these models. One major line of work strives to understand and explain the prediction of a neural network model using the attribution of input words to prediction~\cite{sundar2017,gradientInput}. 

\begin{figure}[]
\begin{center}
\includegraphics[width=\columnwidth]{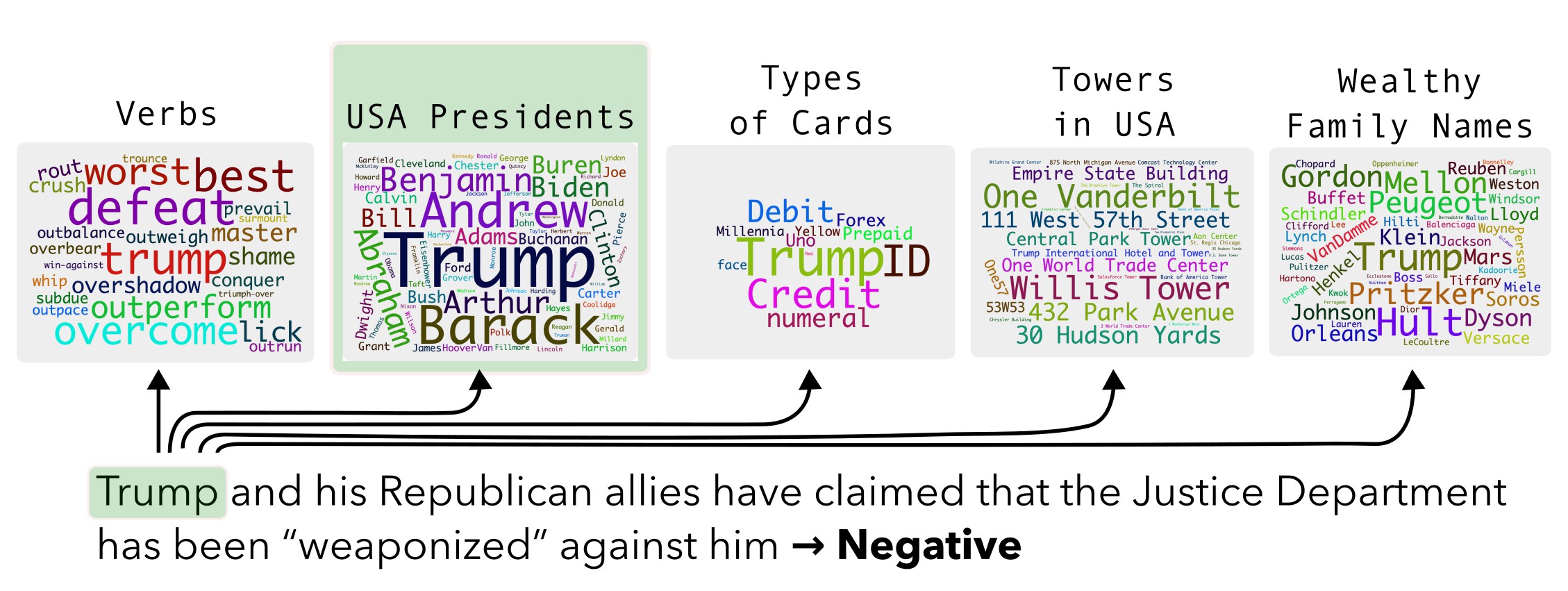}
\end{center}
\vspace{-4mm}
\caption{An example of various facets of word ``trump''}
\vspace{-4mm}
\label{fig:facet_trump}
\end{figure}

However, the explanation based solely on input words is less informative due to the discrete nature of words and the lack of contextual verbosity. A word consists of multifaceted aspects, such as semantic, morphological, and syntactic roles in a sentence. Consider the word ``trump'' in Figure~\ref{fig:facet_trump}. It has several facets such as a verb, a verb with specific semantics, and a named entity 
representing a certain aspect such as tower names,  family names, etc. We argue that given various contexts of a word in the training data, the model learns these diverse facets during training. Given a test instance, depending on the context a word appears, the model uses a particular facet of the input words in making the prediction. The explanation based on salient words alone does not reflect the facets of the word the model has used in the prediction and results in a less informed explanation.

\citet{dalvi2022discovering} showed that the latent space of DNNs represents the multifaceted aspects of words learned during training. The clustering of training data contextualized representations provides access to these multifaceted concepts, hereafter referred to as \emph{latent concepts}. 
%Every layer forms its own latent space and given the contextualized nature of neural network models, a word's position in the latent space depends on the very context it appears in. 
Given an input word in context at test time, we hypothesize that the alignment of its contextualized representation to a latent concept represents the facet of the word being used by the model for that particular input. %Note that the aligned facet is dependent on the context the word is being used in that particular input. 
We further hypothesize that this latent concept serves as a correct and enriched explanation of the input word.
%of a salient input word of a prediction. 
To this end, we propose the
LAtent COncept ATtribution (\LACOAT{}) method that generates an explanation of a model's prediction using the latent concepts.
%of the most salient input features. 
\LACOAT{} discovers latent concepts of every layer of the model by clustering 
%high-dimensional 
contextualized representations of words in the training corpus. Given a test instance, it identifies the most salient input representations of every layer with respect to the prediction and dynamically maps 
%of the salient words 
them to the latent concepts of the training data. The shortlisted latent concepts serve as an explanation of the prediction. 
Lastly, \LACOAT{} integrates a plausibility module that generates 
%takes the latent concept-based explanation as input and generates 
a human-friendly explanation of the latent concept-based explanation.

\LACOAT{} is a local explanation method that provides an explanation of a single test instance. The reliance on the training data latent space makes the explanation reliable and further reflects on the quality of learning of the model and the training data. We perform qualitative and quantitative 
%\textcolor{black}{
evaluation of \LACOAT{} using 
four classification tasks across four 
%the part-of-speech (POS) tagging and sentiment classification tasks across three 
pre-trained models. \LACOAT{} generates an enriched explanation 
%}
%of predictions 
that is useful in understanding the model's reasoning for a prediction. 
%It also helps in understanding how the model has structured the knowledge of a task. 
We also conduct a human evaluation to measure the utility of \LACOAT{} with a human-in-the-loop. 
%\textcolor{black}{
Moreover, we measure the faithfulness of the most salient latent concept to the prediction using representation manipulation and show that it alters
%flips 
the prediction up to 46\% of the time.

\section{Methodology}

%\nd{A figure describing work flow and and an overall summary of the methodology would be good to remind the reader.} \hs{i tried to clarify the pipeline in the section}

%\nd{First module in LACOAT in the pre-processing step and the remaining three are part of the flow i.e. they pull out relevent concept(s) from the data-base for explanation. Do we want to make this kind of distinction in the process?} \hs{i think we can call it as a preprocessing step but the information of pulling from database sounds like implementation details}

\LACOAT{}
%'s pipeline 
consists of the following four modules:
%as shown in Figure \ref{fig:pipeline}. % depicts the various modules and shows the flow of data pictorially. 
%At a high level, 
\begin{itemize}[leftmargin=*,parsep=1pt,topsep=1pt]%[itemsep=2pt,topsep=-2pt]
	\item The first module, \mone{}, discovers latent concepts of a model given a corpus.
	\item \mtwo{}, the second module, selects the most salient words (along with their contextual representations) in a sentence with respect to the model's prediction.
	\item %The third module, 
 Thirdly, \mthree{}, maps the representations of the salient words to the latent concepts 
 %space of the model and provides a concept-based explanation using concepts 
 discovered by \mone{} and provides a latent concept-based explanation.
	\item \mfour{} takes 
 a
 %the 
 latent concept
 %-based 
 explanation as input and generates a plausible and human-understandable explanation of the prediction. 
 
\end{itemize}

%\LACOAT{} works at layer-level. Every layer forms its own latent space and the generated explanation is with respect to the selected layer. 

\vspace{3mm}
Consider a sentiment classification dataset and a sentiment classification model as an example. 
\LACOAT{} works as follows: \mone{} takes the training dataset and the model as input and outputs the model's latent concepts.  At test time, given an input sentence, \mtwo{} identifies the most salient input representations related to the prediction. % at any given layer. 
%For a layer, 
\mthree{} maps these salient input representations to the training data's latent concepts and provides them as an explanation for the prediction. \mfour{} takes the test sentence along with its concept-based explanation to generate a human-friendly and insightful interpretation of the prediction.
%Note that every layer of the model forms its own latent space and the explanation of \LACOAT{} is with respect to a layer.
%
%In the following, we describe these modules in detail. 

Consider $\modelM$ as the DNN model being interpreted, which has $L$ layers, each of size $H$. Let $\overrightarrow{z}_{w_i}$ be the \emph{contextual representation} of a word $w_i$ in an input sentence ${w_1, w_2, ..., w_i, ...}$. The representation can belong to any layer in the model, and \LACOAT{} will generate explanations with respect to that layer.

%depicted by a vector $\overrightarrow{z}^l_{w_i}$
%. $\overrightarrow{z}^l_{w_i}$ is a vector 
%of size $H$ corresponding to the output at layer $l$. %$\overrightarrow{z}^l_{w_i}$ is  and time step $i$ when performing a forward pass through $\modelM$ with the sentence as input. 

\subsection{\mone{}}

%\nd{Although the work is done in Dalvi, but I think we can still motivate it a bit and also align it to this work before going into technical details. }

The words are grouped in the high-dimensional space based on various latent relations such as semantic, morphology, and syntax~\cite{Mikolov:2013:ICLR,bert_geometry:nips2019}. With the inclusion of context, i.e. contextualized representations, these groupings evolve into dynamically formed clusters that represent a unique facet of the words called \emph{latent concept}~\cite{dalvi2022discovering}. %\fd{I think the paragraph can be removed}
%defined a \emph{Concept} as a set of words that are grouped together based on some underlying relationship. Specifically, they 
%called these groupings of words based on their \emph{contextual representations}, derived from some model as \emph{Latent Concepts}. 
%\ym{Remove: 
Figure~\ref{fig:facet_trump} shows a few examples of latent concepts that capture different facets of the word ``trump''.
% Figure \ref{fig:facet_trump} shows a few examples of latent concepts that capture different facets of the word "trump". 
%samples of a few latent concepts learned in a trained BERT model. We follow their methodology to discover latent concepts in any given model, and describe the same below.

The goal of \mone{} is to discover latent concepts given a model $\modelM$ and a dataset $\dataD$. 
%The latent concepts represent how the model has structured the knowledge of the dataset. 
We follow an identical procedure to \citet{dalvi2022discovering} to discover latent concepts. Specifically, for every word $w_i$ in $\dataD$, we extract contextual representations $\overrightarrow{z}_{w_i}$. % from a large corpus of text 
%\nd{specific data used for tuning pLM towards the task} \hs{this would a specific usage and should come in the experimental details}
%
%in order to discover latent concepts learned by a model. This is achieved by first extracting $\overrightarrow{z}^l_{w_i}$ from $\modelM$ for some dataset $\mathcal{D}$, and then 
We then cluster these 
%contextualized 
representations using agglomerative hierarchical clustering~\cite{gowda1978agglomerative}.  
The distance between any two representations is computed using the squared Euclidean distance, and Ward's minimum-variance criterion is used to minimize total within-cluster variance. 
%The algorithm has a hyperparameter $K$ that defines the number of clusters. 
%, so some concepts may have a mixture of word and sentence representations. 
%We optimize $K$ for each dataset as suggested by \citet{dalvi2022discovering}.

Each cluster represents a latent concept. Let $\mathcal{C} = {C_1, C_2, ..., C_K}$ represent the set of latent concepts extracted by \mone{}, where each $C_i = {w_1, w_2, ...}$ is a set of words in a particular context. For sequence classification tasks, we also consider the \texttt{[CLS]} token (or a %model's 
representative classification token) from each sentence in the dataset as a ``word'' and discover the latent concepts. In this case, a latent concept may consist of words only, \texttt{[CLS]} tokens only, or a mix of both. 
%a mix of words and \texttt{[CLS]} tokens, or \texttt{[CLS]} tokens only.

%\subsection{Salient Representations Extraction}
\subsection{\mtwo{}}
Given an input instance $s$, the goal of \mtwo{} is to extract salient input representations with respect to the prediction $p$ from model $\modelM$.
%for some given input. 
Gradient-based methods have been effectively used to compute the saliency of the input features for the given prediction, such as pure Gradient~\cite{Simonyan14a}, Input x Gradient~\cite{learningimportantfeatures} and Integrated Gradients (IG)~\cite{pmlr-v70-sundararajan17a}. 
%For a given input $s$ and prediction $p$, gradient-based methods give attribution scores for each token in the input sequence estimating their importance to the prediction. 
In this work, we use IG as our gradient-based method as it is a well-established method from the literature. However, 
%this module of 
\LACOAT{} is agnostic to the choice of the attribution method, and any other method that identifies salient input representations can be used while keeping the rest of the pipeline unchanged.  Formally, we first use IG to get attribution scores for every token in the input $s$, and then select the top tokens that make up $50\%$ of the total attribution mass (similar to top-P sampling).

\subsection{\mthree}
At test time, given an input sentence 
\mtwo{} provides the salient input representations.
\mthree{} maps each salient representation to a latent concept $C_i$ of the training latent space. These latent concepts highlight a particular facet of the salient representations used by the model and serve as an explanation of the prediction. \mthree{} uses a logistic regression classifier that maps a representation $\overrightarrow{z}_{w_i}$ to one of the $K$ latent concepts.  The model is trained using the representations of words from $\dataD$ that are used by \mone{} as input features and the concept index (cluster id) as their label.   
Hence, for a concept $C_i$ and a word $w_j \in C_i$, a training instance of the classifier is the input $x=\overrightarrow{z}_{w_j}$ and the output is $y=i$.

\subsection{\mfour{}}

Interpreting latent concepts can be challenging due to the need for diverse knowledge, including linguistic, task-specific, worldly, and geographical expertise (as seen in Figure~\ref{fig:facet_trump}). 
\mfour{} offers a user-friendly summary and explanation of the latent concept and its relationship to the input instance using a Large Language Model (LLM). Our intuition of natural language explanation is similar to \citet{singh2023explaining}, however, similar to \citet{mousi2023llms} we relied on latent concepts compared to most activated ngrams. For a given input sentence and the corresponding latent concept, we prompt an LLM to elucidate the relationship between the two. %Due to space constraints, 
%The prompts used for sequence labeling and classification tasks are provided in Appendix \ref{app:prompt}.
We use the following prompt for the sequence classification task:
%sentence classification task:

{
\footnotesize
\begin{verbatim}
Do you find any common semantic, structural, lexi-
cal and topical relation between these sentences 
with the main sentence? Give a more specific and 
concise summary about the most prominent relation
among these sentences.

main sentence: {sentence}
{sentences}
No talk, just go.
\end{verbatim}
}

\noindent and the following prompt for the 
%POS
sequence labeling 
task:
{
\footnotesize
\begin{verbatim}
Do you find any common semantic, structural, lexi-
cal and topical relation between the word highlig-
hted in the sentence (enclosed in [[ ]]) and the 
following list of words? Give a more specific and 
concise summary about the most prominent relation 
among these words.

Sentence: {sentence}
List of words: {words}
Answer concisely and to the point.
\end{verbatim}
}

We did not provide the prediction 
%of the model, 
or the gold label to LLM to avoid biasing the explanation.

\section{Experimental Setup}
\begin{figure*}[]
\centering
\includegraphics[width=0.95\linewidth]{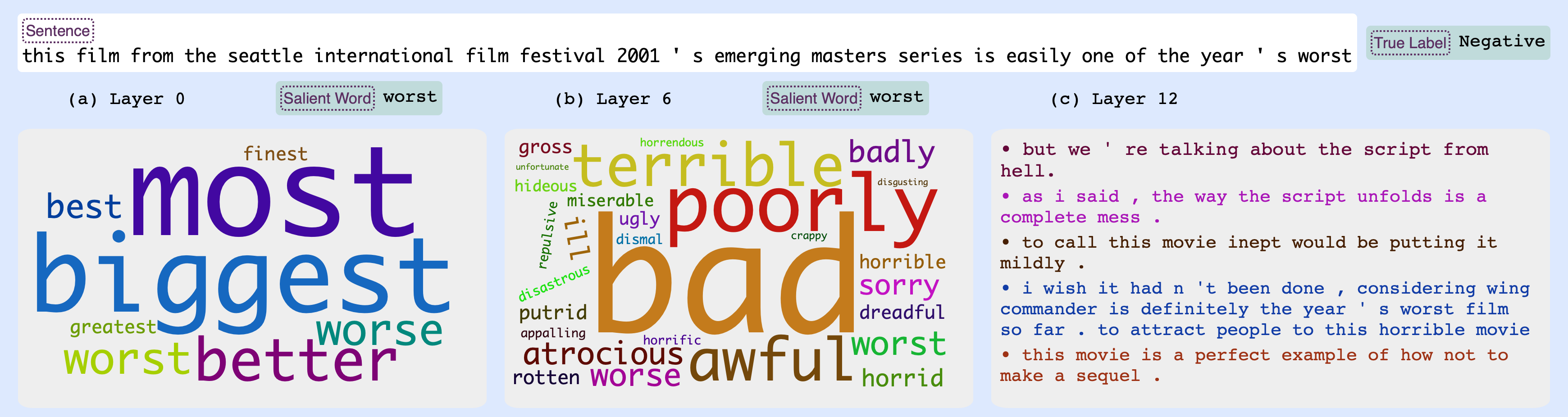}    
\vspace{-3mm}
\caption{Sentiment task: Latent concepts of the most attributed words in Layers 0, 6 and 12}
\vspace{-3mm}
\label{fig:layersieexamples1}
\end{figure*}

% \begin{figure*}[]
% \begin{subfigure}[b]{.99\linewidth}
%     \includegraphics[width=\linewidth]{figures/Example1/concept-layers-1.png}    
% \end{subfigure}
% \begin{subfigure}[b]{.99\linewidth}
%     \includegraphics[width=\linewidth]{figures/Example2/concept-layers-2.png}    
% \end{subfigure}

% \begin{subfigure}[b]{.28\linewidth}
% \includegraphics[width=\linewidth]{figures/Example1/layer0.png}
% \caption{Layer 0}
% \label{subfig:layerwise1:layer0}
% \end{subfigure}
% \begin{subfigure}[b]{.28\linewidth}
% \includegraphics[width=\linewidth]{figures/Example1/layer6.png}
% \caption{Layer 6}
% \label{subfig:layerwise1:layer6}
% \end{subfigure}
% \begin{subfigure}[b]{.28\linewidth}
% \includegraphics[width=\linewidth]{figures/Example1/layer12.png}
% \caption{Layer 12}
% \label{subfig:layerwise1:layer12}
% \end{subfigure}
% \begin{subfigure}[b]{.28\linewidth}
% \includegraphics[width=\linewidth]{figures/Example2/layer0.png}
% \caption{Layer 0}
% \label{subfig:layerwise2:layer0}
% \end{subfigure}
% \begin{subfigure}[b]{.32\linewidth}
% \includegraphics[width=\linewidth]{figures/Example2/layer6.png}
% \caption{Layer 6}
% \label{subfig:layerwise2:layer6}
% \end{subfigure}
% \begin{subfigure}[b]{.28\linewidth}
% \includegraphics[width=\linewidth]{figures/Example2/layer12.png}
% \caption{Layer 12}
% \label{subfig:layerwise2:layer12}
% \end{subfigure}

% \vspace{-2mm}
% \caption{Sentiment task: Examples of the latent concepts of the most attributed words in Layers 0, 6 and 12}
% \vspace{-2mm}
% \label{fig:layersieexamples1}
% \end{figure*}
%\subsection{Procedure and Settings}

\paragraph{Data} 
%\textcolor{black}{
We use Parts-of-Speech (POS) tagging, toxicity classification (Toxicity), sentiment classification (Sentiment), and natural language inference (NLI) tasks for our experiments. POS is a sequence labeling task, while the other tasks are sequence classification tasks. We use the Penn TreeBank dataset~\cite{marcus-etal-1993-building} for POS, the Jigsaw Toxicity dataset~\cite{cjadams_2017} for toxicity, the ERASER Movie Reviews dataset~\cite{eraser_sst} for sentiment, and the MNLI dataset~\cite{wang2019} for the NLI task. Appendix~\ref{app:datasets}
provides information about each dataset.
%Classification. 
%The POS 
%tagging 
%dataset consists of 36k, 1.8k and 1.9k splits for train, dev and test respectively and 44 classes.

%The ERASER movie review dataset consists of labeled paragraphs with human annotations of the words and phrases. We filter sentences that have a word/phrase labeled with sentiment and create a sentence-level sentiment classification dataset. 
%The dataset consists of 9.4k positive and 8.6k negative instances. We randomly split it into 13k, 1.5k and 2.7k splits for train, dev and test respectively. 

\paragraph{Models} We fine-tune 12-layered pre-trained models; BERT-base-cased ~\cite{devlin2018bert}, RoBERTa-base~\cite{liu2019roberta} and XLM-Roberta ~\cite{xlm-roberta} using the training datasets of the 
%two 
tasks considered. 
%\textcolor{black}{
For Llama-2-7b-chat-hf~\cite{touvron2023llama}, we 
use the base model without finetuning with zero-shot prompting for each task.
%}
We use \emph{transformers}~\cite{wolf-etal-2020-transformers} with the default settings and hyperparameters. 
%\textcolor{black}{
Task-wise performance of the models is provided in Appendix (Tables~\ref{tab:pos_dataStats}, ~\ref{tab:eraser_dataStats},~\ref{tab:toxicity_dataStats}, and~\ref{tab:mnli_dataStats}). 

\paragraph{Module-specific hyperparameters} When extracting the activation\footnote{We used the NeuroX toolkit \cite{dalvi-etal-2023-neurox}.} and/or attribution of a word, we average the respective value over the word's subword units \cite{durrani-etal-2019-one}. We optimize the number of clusters $K$ for each dataset as suggested by \citet{dalvi2022discovering}. 
%\textcolor{black}{
We use $K=600$ (POS, Toxicity) and $K=400$ (Sentiment, MNLI) for \mone{}.

Since the number of words in $\dataD$ can be very high, and the clustering algorithm is limited by the 
%total 
number of representations it can efficiently cluster, we filter out words with frequencies less than 5 and randomly select 20 contextual occurrences of every word with the assumption that a word may have a maximum of 20 facets. These settings are in line with \citet{dalvi2022discovering}. In the case of \texttt{[CLS]} tokens, we keep all of the instances.  
%very infrequent ($f_1< 5$) words and randomly select ($f_2 = 10$) contextual representations for every unique word (type) in $\mathcal{D}$.

We use a zero-vector as the baseline vector in \mtwo{}'s IG, using 500 approximation steps. For \mthree{}, we use the cross-entropy loss with L2 regularization and train the classifier with `lbfgs' solver and 100 maximum iterations. 
To optimize the classifier and to evaluate its performance, we split the dataset $\mathcal{D}$ into train ($90\%$) and test ($10\%$).
%and minimize the cross-entropy loss over all the representations. 
\mthree{} used in the \LACOAT{} pipeline is trained using the full dataset $\mathcal{D}$. 
Finally, for \mfour{}, we use ChatGPT 
%as the LLM 
with a \texttt{temperature} of 0 and a \texttt{top\_p} value of 0.95.

\section{Evaluation}
We perform a qualitative evaluation, a human evaluation and a module-level evaluation of \LACOAT{} to measure its correctness and efficacy. 
We find consistent results across all tasks and models. Due to space limitation, we mainly present the results of POS and Sentiment using the BERT and RoBERTa models in the main paper. The full set of results are presented in Apps.~\ref{app:toxicity},~\ref{app:mnli},~\ref{app:llama2}.

\subsection{Qualitative Evaluation}
In this section, we qualitatively evaluate the usefulness of the latent concept-based explanation and the generated human-friendly explanation. 

\subsubsection{Evolution of Concepts}
\LACOAT{} generates the explanation for each layer with respect to a prediction. The layer-wise explanation shows the evolution of concepts in making the prediction. 
Figure~\ref{fig:layersieexamples1} shows layers 0, 6 and 12's latent concept of the most attributed input token for RoBERTa fine-tuned on the sentiment task (see Figure \ref{fig:layersieexamples2} in Appendix for more examples). 
%with respect to the prediction. 
We found that the initial layer 
%(layer 0) 
latent concepts do not always align with the sentiment of the input instance and may represent a general language concept. %correspond to the facet of the word independent of the context. 
For instance, Figure~\ref{fig:layersieexamples1}(a) 
%the latent concept of the first example 
shows the concept of comparative and superlative adjectives of both positive and negative sentiments and is not limited to representing the negative sentiment of the most attributed word. 
% Similarly, Figure~\ref{fig:layersieexamples1}(d) 
% %the latent concept of the second example correspond to
% shows different forms of the verb ``sit'' and is not aligned with its usage in the input instance. 
In the middle layers, the latent concepts evolved into concepts that align better with the sentiment of the input sentence. % as can be seen in Figure~\ref{fig:layersieexamples1}(b). 
% and~\ref{fig:layersieexamples1}(e). 
For instance, the latent concept of Figure~\ref{fig:layersieexamples1}(b) 
%layer 6's latent concept of the first example 
shows a mix of adjectives and adverbs of negative sentiment, i.e. aligned with the sentiment of the input sentence. 
In the sentiment task, the most attributed word in the last layer is \texttt{[CLS]} which resulted in latent concepts consisting of \texttt{[CLS]} representations of the most related sentences to the input. In such cases, we randomly pick five \texttt{[CLS]} instances from the latent concept and show their corresponding sentences in the figure (see Figure~\ref{fig:layersieexamples1}(c)). 
% and~\ref{fig:layersieexamples1}(f)).
We found that the last layer's latent concepts are best aligned with the input instance and its prediction and are the most informative explanation of the prediction. In the rest of the paper, we focus our analysis on the explanations generated using the last layer only and perform a human evaluation to evaluate their efficacy and correctness. 

\begin{figure*}[!htb]
\begin{center}
\begin{subfigure}[!htb]{.40\linewidth}
\includegraphics[width=\linewidth]{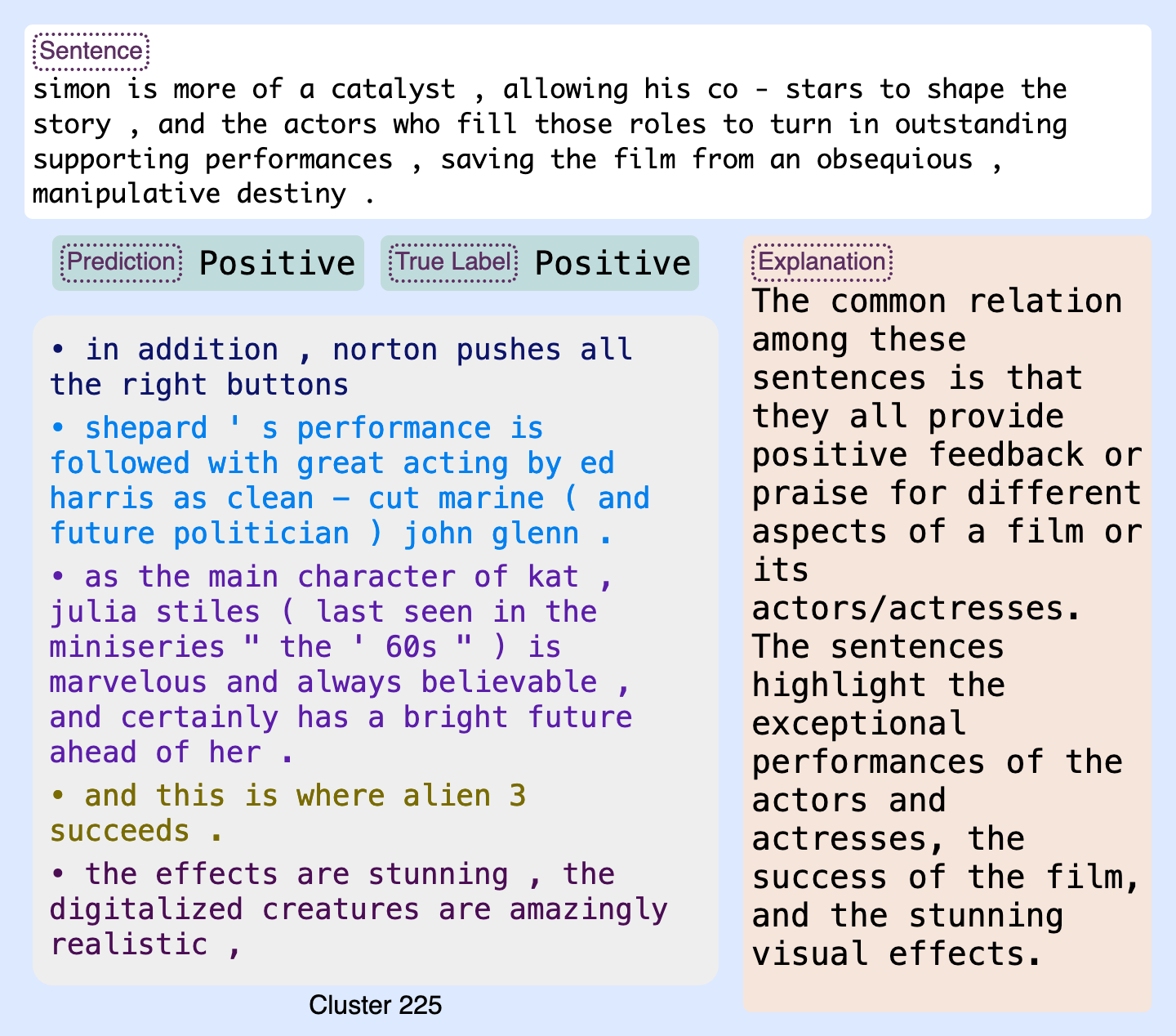}
\caption{Sentiment: A positive labeled test instance correctly predicted by the model.}
\label{subfig:eraser:a}
\end{subfigure}
\begin{subfigure}[!htb]{.45\linewidth}
\includegraphics[width=\linewidth]{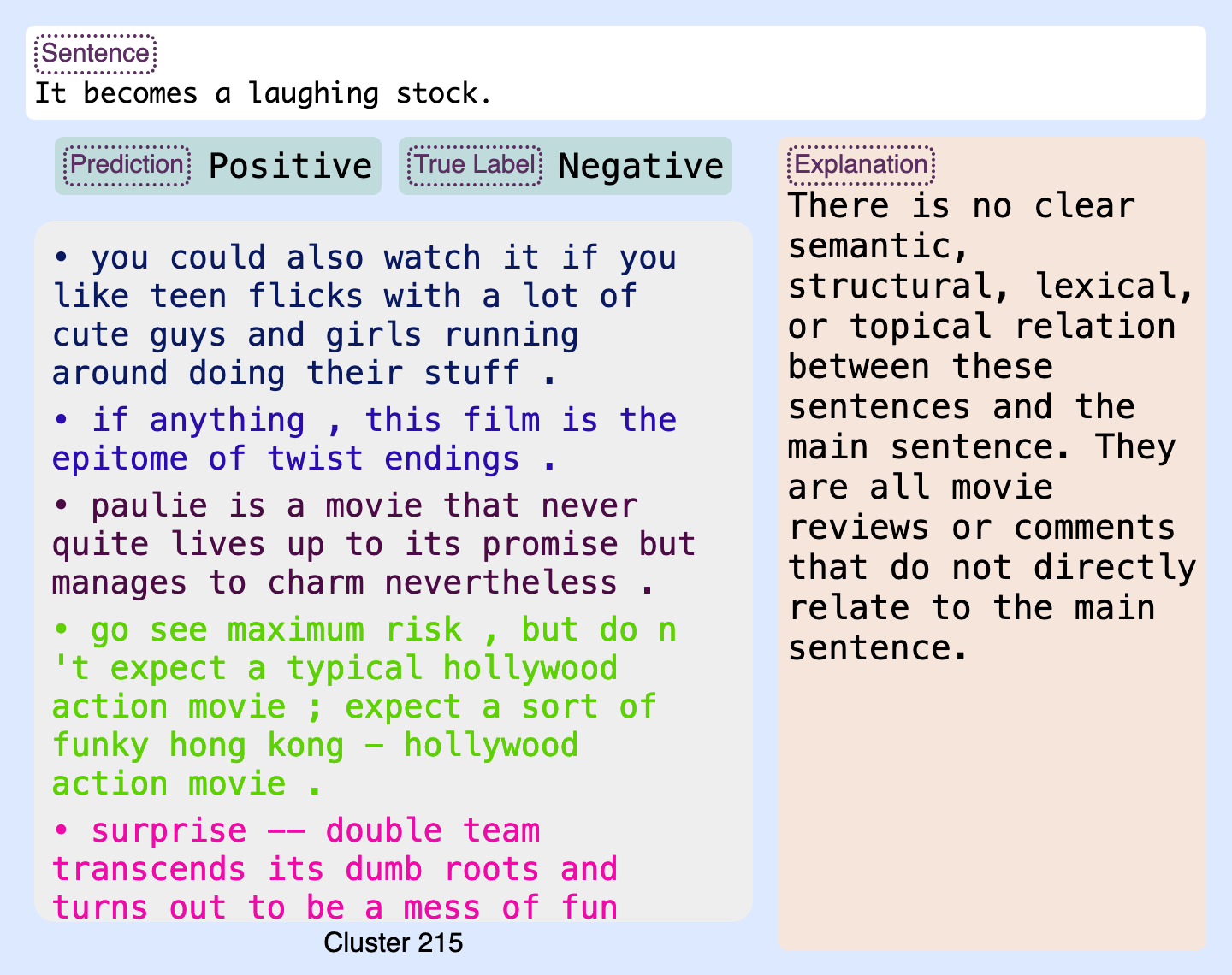}
\caption{Sentiment: A negatively labeled test instance that is incorrectly predicted as positive.}
\label{subfig:eraser:b}
\end{subfigure}
\begin{subfigure}[!htb]{.45\linewidth}
\includegraphics[width=\linewidth]{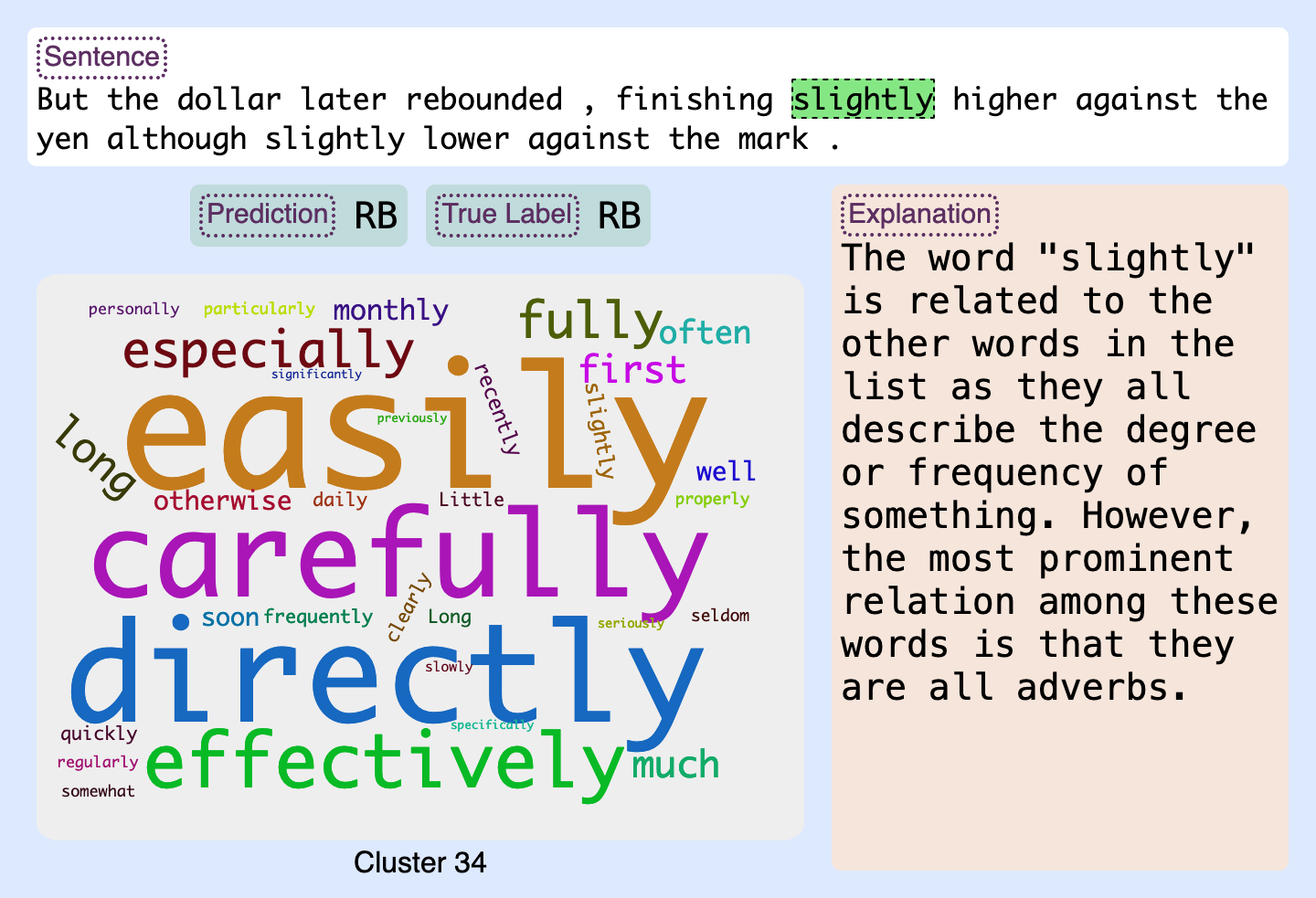}
\subcaption{POS: An adverb with semantics showing degree and intensity of an action}
\label{subfig:a}
\end{subfigure}
\begin{subfigure}[!htb]{.40\linewidth}
\includegraphics[width=\linewidth]{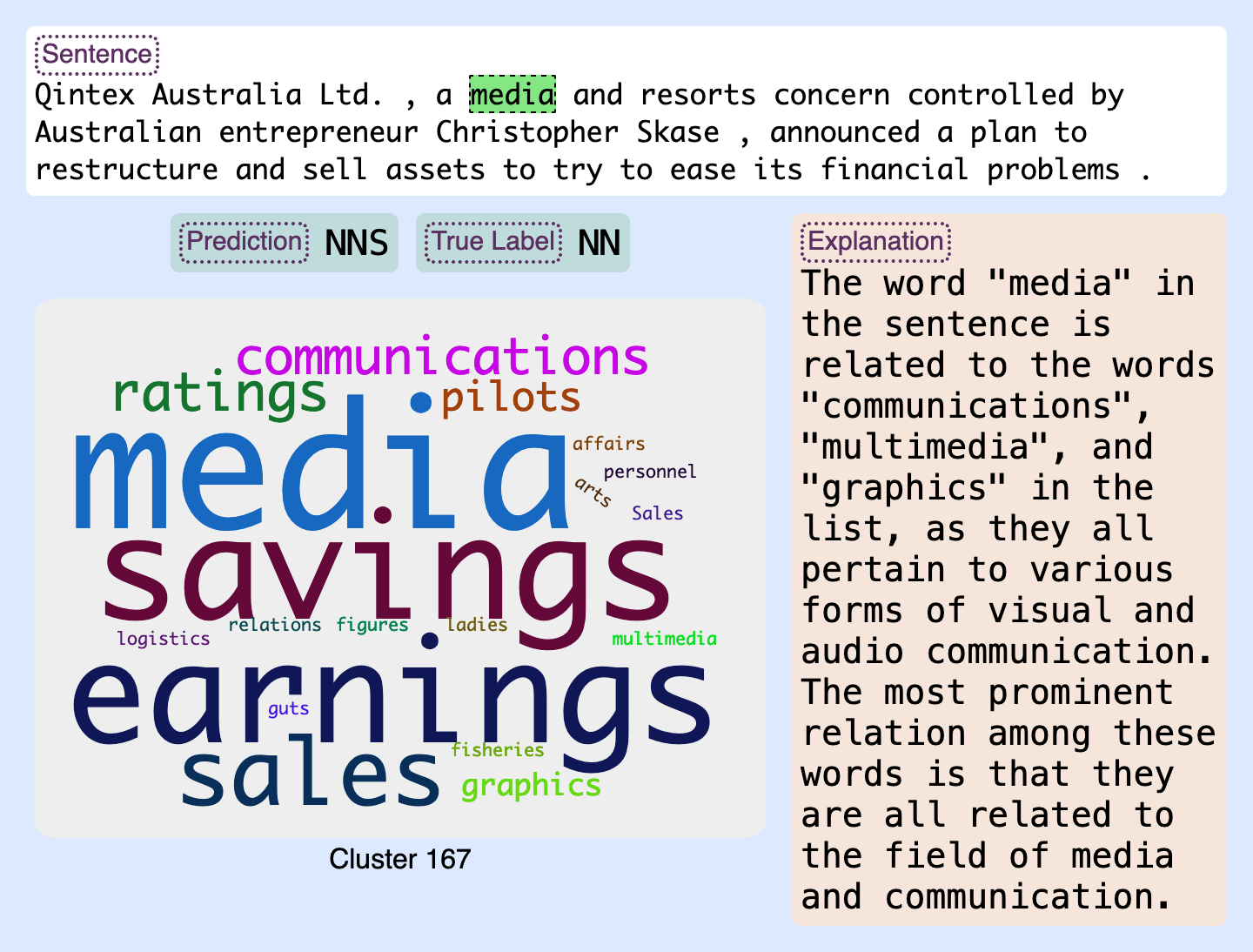}
\subcaption{POS: An incorrect prediction that can be detected from the latent concept}
\label{subfig:b}
\end{subfigure}

\end{center}
%\vspace{-4mm}
\caption{A few examples of \LACOAT{} explanations for BERT using POS and Sentiment tasks}
%\vspace{-4mm}
\label{fig:eraser_pos_examples}
\end{figure*}

\subsubsection{Analyzing Last Layer Explanations}
Figure~\ref{fig:eraser_pos_examples} presents various examples of \LACOAT{} for both POS tagging and sentiment tasks using BERT. The \emph{sentence} refers to the input sentence, \emph{prediction} is the model's output, and \emph{true label} is the gold label. The \emph{explanation} is the final output of \LACOAT. \emph{Cluster X} denotes the latent concept aligned with the most salient word representation at the 12th layer, where X is the cluster ID. For sentiment, we randomly select five \texttt{[CLS]} instances from the latent concept and show their corresponding sentences in the figure.

\paragraph{Correct prediction with correct gold label}
Figures~\ref{subfig:eraser:a} and~\ref{subfig:a} present a case of correct prediction with latent-concept explanation and human-friendly explanation. The former is harder to interpret especially in the case of sentence-level latent concepts as in Figure~\ref{subfig:eraser:a} compared to latent concepts consisting of words (Figure~\ref{subfig:a}). However, in both cases, \mfour{} highlights additional information about the relation between the latent concept and the input sentence. For example, it captures that the adverbs in Figure~\ref{subfig:a} have common semantics of showing degree or frequency. Similarly, it highlights that the reason for the positive sentiment in Figure \ref{subfig:eraser:a} arises from praising different aspects of a film and its actors and actresses.  

%Comparatively, it is easier to The word-level latent concept  explanation highlights that all these sentences "evaluate and praise performances and technical aspects" which are in line with the prediction of the model.  

\paragraph{Wrong prediction with correct gold label}
Figures~\ref{subfig:eraser:b} and \ref{subfig:b} show rather interesting scenarios where the predicted label is wrong. 
In 
%the case of the sentiment classification (
Figure~\ref{subfig:eraser:b},
%), 
the input sentence has a negative sentiment but the model predicted it as positive. The instances of latent concepts show sentences with mixed sentiments such as ``manages to charm'' and ``epitome of twist endings'' is positive, and ``never quite lives up to its promise'' is negative. This provides the domain expert an evidence of a possible wrong prediction. 
%\fd{How? would it be better to say something like "This provides a user with an understanding why the model may have been confused} 
% I am not sure how it let the user understands the cause of the error
The \mfour's \emph{explanation} is even more helpful as it clearly states that ``there is no clear ... relation between these sentences ...". 
%is even more clear in highlighting the difference between the input sentence and the sentence of the latent concept "the main sentence leans more towards a direct negative judgment while the others offer more nuanced or mixed evaluations". This level of explanation is good enough for a domain expert to flag a prediction. 
Similarly, in the case of 
%the 
POS 
%tagging 
%example 
(Figures~\ref{subfig:b}) while the prediction is Noun, the majority of words in the latent concepts are plural Nouns, giving evidence of a possibly wrong prediction. In addition, the \emph{explanation} did not capture any morphological relationship between the concept and the input word.

To study how the explanation would change if it is a correct prediction, we employ 
%the 
TextAttack 
%tool
\cite{morris2020textattack} to create an adversarial example of the sentence in Figure~\ref{subfig:eraser:b} that flips its prediction. The new sentence replaces ``laughing'' with ``kidding'' which has a similar meaning but flipped the prediction to a correct prediction. Figure~\ref{example4} in the Appendix. shows the full explanation of the augmented sentence. With the correct prediction, the latent concept changed and the \textit{explanation} clearly expresses a negative sentiment ``... all express negative opinions and criticisms ...'' compared to the explanation of the wrongly predicted sentence.

%, we can find that the label of the augmented sentence becomes positive, which is matched with the gold label. Furthermore, the new predicted concept is more closely aligned with the main sentence. ChatGPT explained that all sentences express a negative sentiment about the elements in the file such as plots and performance. From the ChatGPT explanations for both original and augmented sentences, the model may not learn the implicit meaning of the 'laughing stock' in the sentence well. Therefore, the model has the wrong prediction for the original sentence.

\begin{figure}[]
\begin{center}
\begin{subfigure}[t]{.9\linewidth}
\includegraphics[width=\columnwidth]{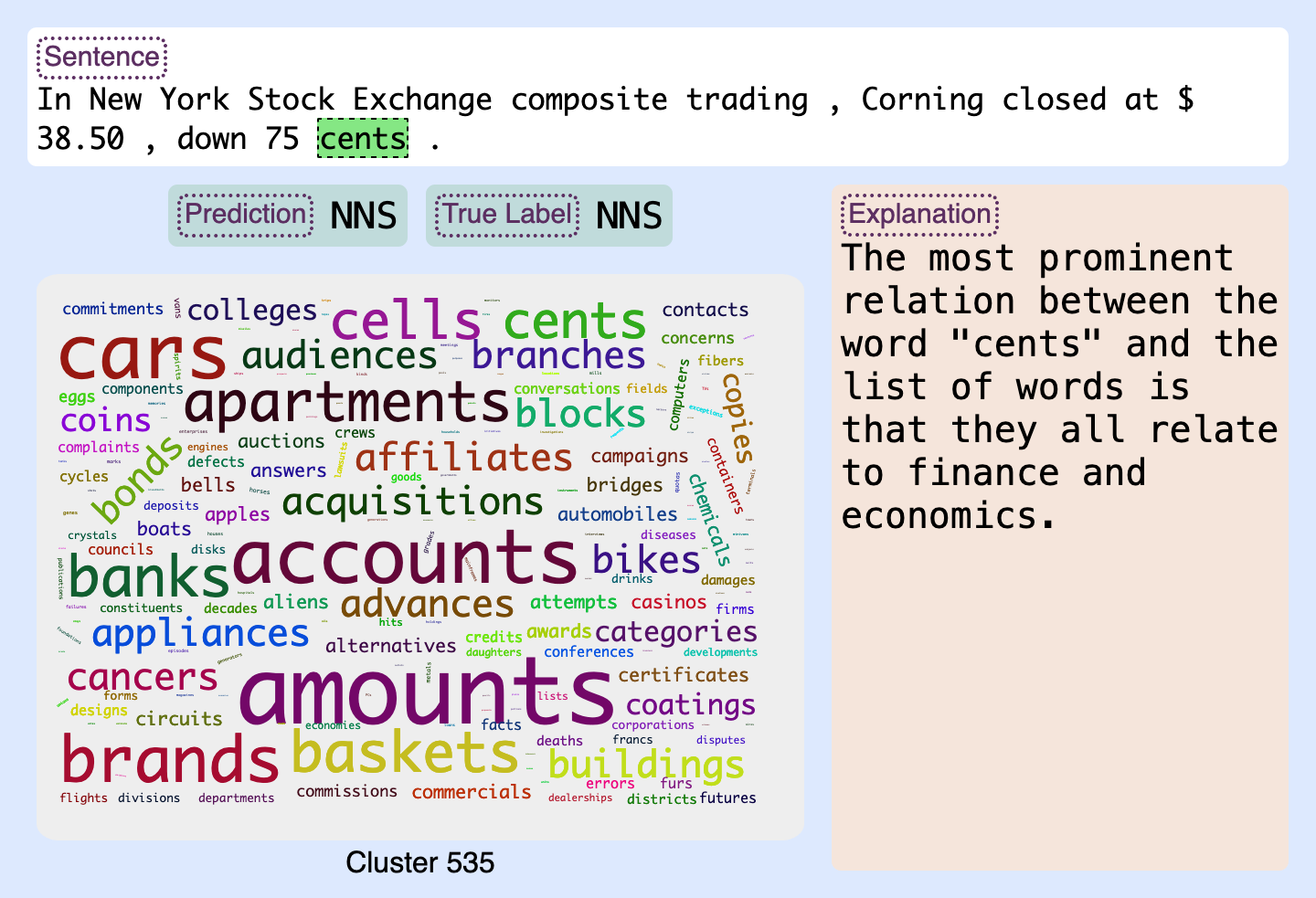}
%\subcaption{RoBERTa}
%\label{subfig:cross_model_a}
\end{subfigure}
\begin{subfigure}[t]{.9\linewidth}
\includegraphics[width=\columnwidth]{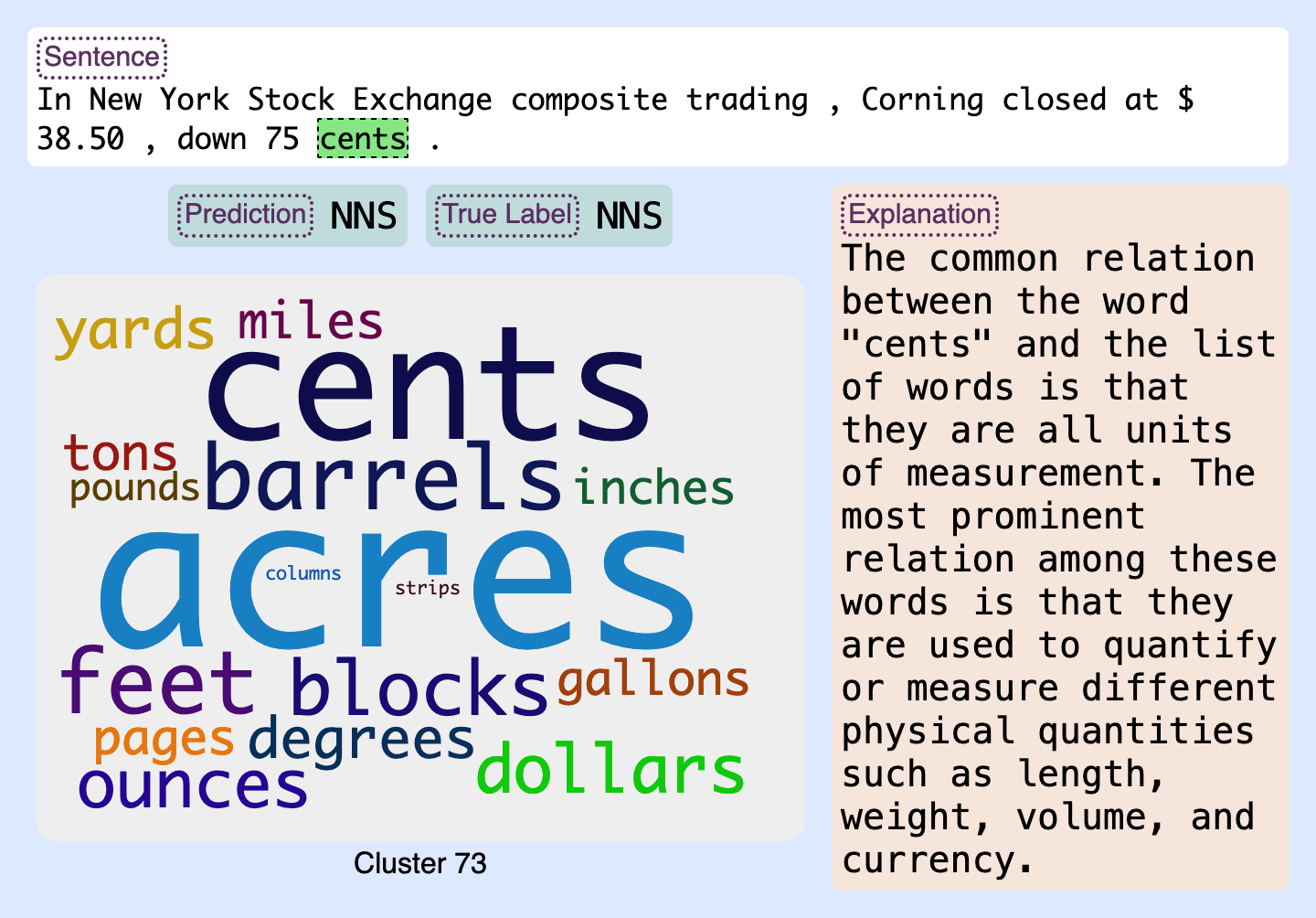}
%\subcaption{XLMR}
%\label{subfig:cross_model_b}
\end{subfigure}
\end{center}
\vspace{-3mm}
\caption{Comparing explanation of XLMR (top) and RoBERTa (bottom)}
\vspace{-4mm}
\label{pos_model_comparison}
\end{figure}

\paragraph{Cross model analysis}
\LACOAT{}
%The latent concept based explanation 
provides an opportunity to compare various models in terms of how they learned and structured the knowledge of a task. Figure~\ref{pos_model_comparison} compares 
%the explanation of 
XLMR (top) and RoBERTa (bottom) for identical inputs. Both models predicted the correct label. However, their latent concept based explanation is substantially different. XLMR's explanation shows a large and diverse concept where many words are related to finance and economics. RoBERTa's latent concept is rather a small focused concept where the majority of tokens are units of measurement. It is worth noting that both models are fine-tuned on identical data.

\subsection{Human Evaluation}
%\textcolor{black}{
We perform two human evaluations; one aimed at evaluating the usefulness of \LACOAT{}'s explanation in understanding a prediction (\LACOAT{} Effectiveness) and the other compares \LACOAT{} with other explanation methods. 
%}
\paragraph{\LACOAT{} Effectiveness}
%To evaluate the explanation generated by \LACOAT{} empirically, 
We conduct a human evaluation using four annotators across 50 test samples. Specifically, given an explanation (e.g. Figure \ref{fig:eraser_pos_examples}), all annotators are asked to answer five questions (Q1-Q5) that aimed at evaluating the usefulness of \LACOAT{}.\footnote{We provide the evaluation questions in Appendix~\ref{app:humanevaluation}.} Specifically,   
\textit{Q1} evaluates whether \LACOAT{} attributes the correct concept to a given prediction, while \textit{Q2} and \textit{Q3} measure the efficacy of \LACOAT{}'s output in helping a user understand the prediction. \textit{Q4} and \textit{Q5} evaluates the output of \mfour{}. They specifically separate out the cases where the explanation was accurate but irrelevant to the task at hand.

Table \ref{tab:human-eval} shows the consolidated labels by picking the majority label in case of Yes/No questions and averaging the annotations in case of the rest. The evaluation shows that the latent concept itself was not only relevant to the task at hand, but also helped the user understand the model's prediction. The results for the helpfulness of the explanation text were 
%more 
mixed, with the majority of the annotations stating that it did not help or hinder their process. 
However 
upon 
%further 
inspection, we see that the explanation 
%text 
was mostly helpful in all the cases where the model made the correct prediction, and not helpful when the prediction was incorrect. Qualitatively analyzing the explanation text for incorrect prediction shows that \mfour{} mostly outputs ``There is no relationship between the sentences and the concepts'', which was deemed as hindering by most of the annotators. While such an explanation may serve as an indicator of a potential problem in the prediction, improving the prompt may result in a response that is indicative of the issue with the prediction. 
%Iterating on the prompt and improving it may help shift the consensus to the helpful side, and 
We leave this exploration for the future. Table~\ref{tab:human-eval} also shows the agreement between the annotators using Fleiss' Kappa.
%scores. 
Since not all samples were annotated by all annotators, we compute the average Fleiss' kappa of each annotator with the consolidated annotation. %In general, 
The agreement ranges from \textit{Fair} to \textit{Substantial} across the five questions.

\paragraph{Comparison with other Methods}
Despite the numerous proposed explanation methods comparing them due to the difference in granularity, type of explanations, and 
methodology. 
%Nevertheless, 
We design a human evaluation where evaluators assign a score between 1 to 3 to each of three explanations generated by IG, \LACOAT{} and Cockatiel~\cite{jourdan_cockatiel_2023}.
%
% In Figure~\ref{fig:layersieexamples1}, layer 0 serves as a direct comparison to the explanation provided by the attribution methods. Here, IG highlights the input word ``worst" in the given
% %and ``stands" for the first and second examples.
% example. While it informs the user of the most salient input token that effects the prediction, it does not provide insights into the particular facet the model has used in that particular example. Even if we compare the explanation of other layers like layer 6, the latent concept of the attributed word is more informative than the attributed word alone.
%We perform human evaluation comparing \LACOAT{} explanations from layer 0, layer 6 and layer 12 with IG and Cockatiel~\cite{jourdan_cockatiel_2023}. Given a test instance, its prediction and the gold label, the evaluators are provided with explanations generated by all three methods and are asked to rank them between 1 to 3 
%where 1 implies useful and 3 implies not useful. 
The annotation setup allows for ranking multiple methods with the same usefulness rating. A total of 400 annotations were collected using four evaluators, each evaluating all test instances. We provide the details of the evaluation setup and the results in Appendix~\ref{app:humanevaluation}. 
The second part of Table~\ref{tab:human-eval} shows the percentage of samples for which each annotator ranked \LACOAT{} as the same or better than both IG and Cockatiel. 
The consolidated ranking is computed by averaging the ranks across users. 
The average Cohen's $\kappa$ indicates \textit{Fair agreement} between each annotator and the consolidated ranking, demonstrating that the \LACOAT{} explanation is more useful for understanding predictions compared to other methods.

\begin{table}[]
\resizebox{\linewidth}{!}{
\centering
\begin{tabular}{@{}l|l|cccr@{}}
\toprule
Top   & \multicolumn{1}{c|}{\multirow{2}{1cm}{Labels}} & \multicolumn{1}{c}{\multirow{2}{1cm}{Correct Samples}} & \multicolumn{1}{c}{\multirow{2}{1cm}{Incorrect Samples}} & \multicolumn{2}{c}{All Samples}           \\ \cmidrule(l){5-6} 
   & \multicolumn{1}{c|}{}                        & \multicolumn{1}{c}{}                                 & \multicolumn{1}{c}{}                                   & Annotation & \multicolumn{1}{l}{Fleiss $\kappa$} \\ \midrule
Q1 & Yes/No                                       & 28 / 0                                                 & 20 / 2                                                   & 48 / 2       & 0.35                     \\
Q2 & Helps/Neutral/Hinders                        & 27 / 1 / 0                                               & 17 / 5 / 0                                                 & 44 / 6 / 0     & 0.41                     \\
Q3 & Helps/Neutral/Hinders                        & 16 / 10 / 2                                              & 1 / 19 / 2                                                 & 17 / 29 / 4    & 0.61                     \\
Q4 & Yes/No                                       & 17 / 11                                                & 5 / 17                                                   & 22 / 28      & 0.47                     \\
Q5 & Yes/No                                       & 17 / 11                                                & 6 / 16                                                   & 23 / 27      & 0.80                     \\ \bottomrule
\end{tabular}
}
\resizebox{\linewidth}{!}{
\centering
\begin{tabular}{@{}lllllll@{}}
\toprule
Bottom                          & A1              & A2              & A3              & A4              & Consolidated             & Average Cohen's $\kappa$             \\ \midrule
LACOAT $\uparrow$ & \multicolumn{1}{r}{85\%} & \multicolumn{1}{r}{72\%} & \multicolumn{1}{r}{77\%} & \multicolumn{1}{r}{87\%} & \multicolumn{1}{r}{89\%} & \multicolumn{1}{r}{0.37} \\ \bottomrule
% LACOAT $\uparrow$ & \multicolumn{1}{r}{82\%} & \multicolumn{1}{r}{72\%} & \multicolumn{1}{r}{72\%} & \multicolumn{1}{r}{78\%} & \multicolumn{1}{r}{88\%} & \multicolumn{1}{r}{0.33} \\ \bottomrule
\end{tabular}
}

\vspace{-2mm}
\caption{\textbf{Top}: Consolidated label distribution for Q1-Q5. Fleiss' $\kappa$ scores are computed by averaging each annotator with the consolidated annotation. The consolidated labels and agreement scores are shown for all the samples, as well as partitioned into those where the model prediction was correct/incorrect. \textbf{Bottom}: Percentage of samples where \LACOAT{} is ranked similar or better than other methods. A$^*$ represents the average preference of \LACOAT{} per annotator.}
\label{tab:human-eval}
\end{table}

% \begin{table}[]
% \centering
% \footnotesize
% \begin{tabular}{r|l|c}
% \toprule
%  & Consolidated label & Fleiss' \kappa                              \\
% \midrule
% Q1       & \textbf{Yes: 48}, No: 2                                     & 0.35                  \\
% Q2       & \textbf{Helps: 44}, Neutral: 6, Hinders: 0  & 0.41                  \\
% Q3       & Helps: 17, \textbf{Neutral: 29}, Hinders: 4 & 0.61                  \\
% Q4       & Yes: 22, \textbf{No: 28}                                    & 0.47                  \\
% Q5       & Yes: 19, \textbf{No: 31}                                    & 0.60                  \\
% Q6       & Yes: 23, \textbf{No: 27}                                    & 0.80          \\
% \bottomrule
% \end{tabular}
% \caption{Consolidated label distribution for Q1-Q6. Fleiss' $\kappa$ scores are computing by average of each annotator with the consolidated annotation.}
% \label{tab:human-eval}
% \end{table}

\subsection{Module Specific Evaluation}

\begin{table}[!t]
\footnotesize
% \vspace{-2mm}
\resizebox{\linewidth}{!}{
\centering
\begin{tabular}{l|cc|ccc}
\toprule
       & \multicolumn{2}{c}{POS} & \multicolumn{3}{|c}{Sentiment} \\
%       & \multicolumn{3}{c}{Saliency based} &  \multicolumn{3}{c}{Saliency based} \\
Layers & BERT     & RoBERTa   & BERT    & RoBERTa & \textcolor{black}{Llama2}    \\
\midrule
9 \textcolor{black}{(/20)}      & 92.38    & 86.97     & 31.94   & 99.59  & 70.63      \\
10 \textcolor{black}{(/24)}    & 92.79    & 89.64     & 99.57   & 99.69  & 75.64   \\
11 \textcolor{black}{(/28)}    & 93.39    & 89.95     & 99.71   & 99.48  & 71.30  \\
12 \textcolor{black}{(/32)}    & 93.95    & 90.04     & 99.25   & 99.27  & 71.02   \\
\bottomrule
\end{tabular}
}
\vspace{-3mm}
\caption{Accuracy of \mtwo{} in mapping a representation to the correct latent concept. The layer number in the parentheses corresponds to the Llama2-chat model.
%See App. Table~\ref{tab:pos_predicted_attribution},~\ref{tab:eraser_predicted_attribution_position},~\ref{tab:eraser_predicted_attribution_saliency} for complete results.
}
\vspace{-4mm}
\label{tab:conceptmapper_eval_short}
\end{table}

%\hs{need to make it general and no specific to POS}
%\fd{Readability will be significantly better by using subsubsections instead of paragraphs, its difficult to see the delineations currently}
%Ideally, we would like to have gold annotations of the latent concepts that are used by the model and we compare them with our methodology to measure its correctness. However, due to the blackbox nature of the model, we do not have this privilege. We design constrainted scenarios to validate different modules of our pipeline. 

%\nd{I skimmed through Section 3.2. I think the content is good but we need to change the terminology. In the opening paragraph, we don't seem to be confident about our own methodology and say that we will make assumptions. Then we are using words such as "Hypothesis". Each result is validating that our methodology for that module is correct.  } \hs{I attempted to improve it}

The correctness of \LACOAT{} depends on the performance of each module it comprised off. 
%For example, are the latent concepts a true representation of the latent space?
The ideal way to evaluate the efficacy of these modules is to consider gold annotations.
However, 
they
%ground truth annotations 
are not available for any module. To mitigate this limitation, 
%of evaluation, 
%and to evaluate each module of \LACOAT, 
we design various constrained scenarios 
%using the POS tagging task 
where certain assumptions can be made about the representations of the model.  
%due to the nature of the task 
%and makes POS tagging well suited for the purpose of evaluation. 
For example, the POS 
%tagging 
model optimizes POS tags so it is highly probable that the last layer representations form latent concepts that are a good representation of POS tags as suggested by various previous works~\cite{kovaleva-etal-2019-revealing,durrani-etal-2022-transformation}. One can assume that for \mone{}, the last layer latent concepts will form groupings of words based on specific tags and for \mtwo{}, the input word at the position of the predicted tag should reside in a latent concept that is dominated by the words with the same tag. In the following, we evaluate the correctness of these assumptions. % in the following subsections. %In the following, we evaluate each module of \LACOAT{}. % using the POS tagging task. Only in certain cases when it is possible to make an observation using the sentiment classification dataset, we present its results as well.

\textbf{Latent Concept Annotation}
\label{text:annotation}
For the sake of evaluation, we annotated the latent concepts automatically using the class labels of each task. Given a latent concept, we annotate it with a certain class if more than 90\% of the words in the latent concept belong to that class. In the case of POS, the latent concepts will be labeled with one of the 44 tags. %In the case of 
For sentiment, the class labels, \textit{Positive} and \textit{Negative}, are at sentence level. We tag a latent concept as \textit{Positive}/\textit{Negative} if 90\% of its tokens (\texttt{[CLS]} or words) belong to sentences labeled as \textit{Positive}/\textit{Negative} in the training data. 
%We follow the same procedure for the \textit{Negative} class. 
The latent concepts that do not fulfill the criteria of 90\% for any class are annotated as \textit{Mixed}.

\subsubsection{\mone{}}
%\textbf{A latent concept is a true reflection of the properties that a representation possesses.} \fd{Would it be better to pose this as a question rather than statement? This will also save a line in the coming sentences}
\mone{} identifies latent concepts by clustering the representation. % in the high dimensional space. 
%These latent concepts reflect the properties by which representations are grouped together. 
We question whether the discovered latent concepts are a true reflection of the properties that a representation possesses. 
%
%Given a sequence of words, the POS tagging model predicts a tag for every word. We hypothesize that the latent concepts on the last layer of the POS tagging must exhibit properties of individual POS tags. 
Using \mone, we form latent concepts of the last layer 
%using the POS training data 
and automatically annotate them as described above.
%We automatically annotate each latent concept with a POS tag on the basis of if 95\% of the words in the latent concept comprise a specific tag. 
We found 87\%, 83\% and 86\% of the latent concepts of BERT, RoBERTa and XLMR that perfectly map to a POS tag respectively. We further analyzed other concepts where 90\% of the words did not belong to a single tag. We found them to be of compositional nature i.e. a concept consisting of related semantics like a mix of adjectives and proper nouns about countries such as Swedish and Sweden (Appendix:Figure~\ref{fig:clusters}). For sentiment, we found 78\%, 95\%, 94\%, \textcolor{black}{and 67\%} of the latent concepts of BERT, RoBERTa, XLMR, and \textcolor{black}{Llama2-chat} to consist of either Positive or Negative sentences. The high number of class-based clusters of RoBERTa and XLMR show that at the 
last
%12th 
layer, the majority of their latent space is separated based on these two classes (see Table~\ref{app:tab:cluster_tag_numbers_eraser} and Table~\ref{app:tab:cluster_tag_numbers_llama_sentiment} for detailed results). %presents these figures for each layer.  

\begin{table}[]
\footnotesize
% \vspace{-2mm}
\resizebox{\linewidth}{!}{
\centering
\begin{tabular}{p{1.1cm}|l|rrrrr}
\toprule
\multicolumn{2}{l}{\textcolor{black}{Layers (BERT)} } & 0   & 2   & 5    & 10  & 12    \\
\midrule
POS    & Top 1 & 100 & 100 & 99.03 & 92.67 & 84.19 \\
       & Top 2 & 100 & 100 & 99.75 & 97.89 & 94.15 \\
       & Top 5 & 100 & 100 & 99.94 & 99.68 & 99.05 \\
\midrule
Sentiment & Top 1 & 100 & 100 & 97.19 & 83.09 & 68.24 \\
       & Top 2 & 100 & 100 & 99.63 & 92.67 & 83.24 \\
       & Top 5 & 100 & 100 & 99.94 & 97.75 & 94.24 \\
\midrule
\multicolumn{2}{l}{\textcolor{black}{Layers (Llama2-chat)}} & 0   & 8   & 16   & 24  & 32    \\
\midrule
Sentiment & Top 1 & 97.55 & 49.46 & 61.86 & 63.95 & 66.47 \\
       & Top 2 & 97.55 & 68.06 & 80.97 & 82.26 & 82.84 \\
       & Top 5 & 97.55 & 86.37 & 95.03 & 94.23 & 94.88 \\
\bottomrule
\end{tabular}
}
\vspace{-3mm}
\caption{
%Top 1, 2 and 5 a
BERT \textcolor{black}{\& Llama2-chat}: Accuracy of \mthree{} in mapping a representation to the correct latent concept. See Table \ref{tab:pos_mapper_eval},~\ref{tab:eraser_mapper_eval} and~\ref{tab:sentimemt_mapper_eval_llama} in the Appendix for detailed results. 
}
\vspace{-4mm}
\label{tab:mapper_eval_bert_pos_eraser}
\end{table}

\subsubsection{\mtwo{}} 
We question whether the salient input representation correctly represents the latent space of the output. This specifically evaluates \mtwo{}. We calculate the number of times the representation of the most salient word/\texttt{[CLS]} token maps to the latent concept of the identical label as that of the prediction. We expect a high alignment at the top layers for \mtwo{} to be correct. We do not include \mthree{} when evaluating this and conduct the experiment using the training data only where we already know the alignment of a salient representation and the latent concept. \textcolor{black}{Table~\ref{tab:conceptmapper_eval_short} shows the results from the last four layers of 12-layered models, as well as the latter four layers of the Llama2-chat model (See Appendix Tables~\ref{tab:pos_predicted_attribution_saliency},~\ref{tab:eraser_predicted_attribution_saliency}, and~\ref{tab:eraser_predicted_attribution_position_llama} for detailed results).} For POS, 
%tagging task, 
% the salient representation is identical for both the position based and saliency based methods and results in the same performance. 
we observed a successful match of 
above 90\% for all models. 
%93.95\%, 90.04\% and 93.13\% for BERT, RoBERTa and XLMR 
%models 
%respectively. 
We observed the mismatched cases and found them to be also of a compositional nature i.e. latent concepts comprised of semantically related words (see Appendix:Figure~\ref{fig:clusters} for examples).

% \ym{Appendix: Table 7(tab:pos\_predicted\_attribution) for POS, Table 8(tab:eraser\_predicted\_attribution\_position) and Table 9(tab:eraser\_predicted\_attribution\_saliency) for ERASER}
% \hs{what is the reason of low results of POS? composition?}
% \hs{add compositional clusters in appendix}
% \hs{update POS numbers with 90\%}

For sentiment, more than 99\% of the time, the last layer's salient representation maps to the predicted class label in  \textcolor{black}{12-layered} models, confirming the correctness of \mtwo. The performance drop for the lower layer is due to the absence of class-based latent concepts in the lower layers i.e. concepts that comprised more than 90\% of the tokens belonging to sentences of one of the classes. \textcolor{black}{By comparison, the Llama2-chat model performs worse than the 12-layer model; however, it still achieves over 70\% accuracy in its last layer}.
%\ym{updated: move position based attribution analysis to appendix}

% The other reason is the position-based method which fails to find the right latent concept when the most attributed word is different from the position of the output head.

%Comparing Position based and Saliency based methods, it can be observed that for the last two layers the results are quite comparable. However, when the most attributed word is other than the position of the output head, the performance of position based method dropped to zero.  from the ERASER results that the latter method is a better choice for identifying the salient representation 
%\hs{why IG is not zero here? is it because that the attribution moved from position based to other tokens?}

%\hs{what is the reason of identical numbers of both methods? since we are picking the top IG, it always attributed to the input word. } 
%We perform a similar experiment using a sentiment classification model under the assumption that the output representation should align with a latent concept whose polarity is identical to the output class. We observed consistent results (see Appendix XYZ***).

\subsubsection{\mthree{}} 
%\textbf{\mthree{} correctly maps a new representation to the latent space.}\fd{If we change the others lets make this as a question as well} 
%Here, w
We evaluate the correctness of \mthree{} in mapping a test representation to the training data latent concepts. 
\mthree{} trains using representations 
%in latent concepts 
and their cluster IDs as labels. 
%For every layer, w
We randomly split this training data 
%of latent concepts, representations and their cluster id 
into 90\% train and 10\% test where 
%data. 
%Here, 
the 
test 
data 
serves as the gold 
%standard 
annotation of latent concepts. We train \mthree{} using the train instances and measure the accuracy of the test instances. \textcolor{black}{Table~\ref{tab:mapper_eval_bert_pos_eraser} 
presents the accuracy of POS using BERT, as well as for the Sentiment tasks using both BERT and Llama2-chat models (See Appendix Tables~\ref{tab:pos_mapper_eval},~\ref{tab:eraser_mapper_eval},~\ref{tab:sentimemt_mapper_eval_llama}).} 
%for all results).}
%of other models). %Top-1, Top-2 and Top-5 refer to top 1, 2 and 5 
%\fd{I think this sentence can be removed}
%predictions of the mapper. 
Observing Top-1 accuracy, the performance of \mthree{} starts high (100\%) for lower layers and drops to 84.19\% and 68.24\% for the last layer \textcolor{black}{of BERT. A similar trend is observed for Llama2-chat, which exhibits 97.55\% accuracy in the lower layer and decreases to 66.47\% in the last layer.} We found that the latent space becomes dense on the last layer. This is in line with \citet{ethayarajh-2019-contextual} who 
%made a similar observation and 
showed that the representations of higher layers are highly anisotropic. This causes 
%several similar 
concepts to be close in the space. 
%We manually tested these latent concepts and found them to be related. 
If true, the correct label should be among the top predictions of the mapper. We empirically tested it by considering the top two and top five predictions of the mapper, achieving a performance of up to \textcolor{black}{99.05\% (BERT) for POS, 94.24\% (BERT) and 94.88\% (Llama2-chat) for Sentiment}.

\subsection{Faithfulness Evaluation}
\label{sec:sec:faithfulness}
% \citet{zhao-aletras-2023-incorporating-faithfulness} proposed masking parts of input token representations to evaluate faithfulness. We adapted their methodology to the latent concept faithfulness evaluation. We consider a salient latent concept highlighted by \LACOAT{} to be faithful to the prediction if the ablation of that latent concept causes a change in prediction performance. 
% %and prediction confidence of the originally predicted class. 
% We define ablation of a latent concept as removing the information of that latent concept from the prediction vector i.e. \texttt{[CLS]}. We calculate the vector of a latent concept by averaging the training representations that belong to the latent concept. At inference time, we subtract the latent concept vector from the \texttt{[CLS]}  representation of layer 12 and perform the prediction. We report the accuracy of the model and the percentage of predictions altered (see Table~\ref{tab:manipulation_roberta} in App.). Moreover, we report the manipulation of \texttt{[CLS]}  using random vectors. The results show a substantial change in all metrics when the latent concept is ablated compared to random, confirming the faithfulness of the latent concept based explanation.

%\citet{zhao-aletras-2023-incorporating-faithfulness} proposed masking parts of input token representations to evaluate faithfulness. We adapted their methodology to the latent concept faithfulness evaluation. 
We consider a salient latent concept highlighted by \LACOAT{} to be faithful to the prediction if the ablation of that latent concept causes a change in prediction performance. 
%and prediction confidence of the originally predicted class. 
We define ablation of a latent concept as removing the information of that latent concept from the prediction vector i.e. \texttt{[CLS]}. We calculate the vector of a latent concept by averaging the training representations that belong to the latent concept. At inference time, \textcolor{black}{we ablate the most salient latent concept for a prediction by subtracting the latent concept vector from the \texttt{[CLS]} representation of the last layer and then perform the prediction. The prediction change represents the percentage of predictions that were altered after the manipulation. As a baseline, we manipulate using a random vector of the same magnitude to that of the salient latent concept vector and report the change in prediction.
%We also aim to evaluate faithfulness by comparing the effect of manipulating a random vector to that of the salient latent concept vector. We generate the random vector with the same magnitude as the salient latent concept vector.
}

\textcolor{black}{Table~\ref{tab:manipulation_roberta} reports the model's accuracy and the percentage of altered predictions across all metrics. The results show that manipulating the \texttt{[CLS]} token representation using the LACOAT vector leads to significant performance drops and prediction changes across all datasets. In contrast, random vector manipulations have minimal impact on the model's performance and predictions. These findings suggest that the LACOAT vector plays a crucial role in the model's decision-making process. Comparing the results across different datasets, MNLI shows a relatively smaller drop in accuracy when manipulating the salient latent concept vector. We suspect this is due to the nature of the MNLI task, which requires reasoning over multiple sentences, with relevant information spread across multiple latent concepts. Nevertheless, the difference in results between the original accuracy and random vector confirms our hypothesis about the faithfulness of latent concepts.}

\begin{table}
\footnotesize
\vspace{-2mm}
\centering
\begin{tabular}{ll|cc}
    \toprule
    \multicolumn{2}{c|}{} & \multicolumn{2}{c}{\textbf{Faithfulness Metrics}} \\
    \midrule
    \textbf{Dataset} & \textbf{Setting} & \textbf{Accuracy} & \textbf{\% Label Flip} \\
    \midrule
    \multirow{3}{*}{\textbf{Sentiment}} & \textbf{Original}  & 96.31 & - \\
    & LACOAT  & 55.91 & 43.98 \\
    & Random  & 96.09 & 0.14 \\
    %& Random_{LC} & 89.24 & \\
    \midrule
    \multirow{3}{*}{\textbf{Toxicity}} & \textbf{Original}  & 91.55 & - \\
    & LACOAT  & 51.78 & 46.44 \\
    & Random  & 91.93 & 0.13 \\
    %& Random_{LC} & 86.26 & \\    
    \midrule
    \multirow{3}{*}{\textbf{MNLI}} & \textbf{Original}  & 87.69 & - \\
    & LACOAT  & 82.08 & 8.83 \\
    & Random  & 88.12 & 0.55 \\
    %& Random_{LC} & 80.42 & \\    
    \bottomrule
\end{tabular}
\caption{\textcolor{black}{Faithfulness evaluation using the RoBERTa model. \textit{Original} refers to the model's performance without any manipulation, \textit{LACOAT} represents the performance after subtracting the most salient latent concept vector from the \texttt{[CLS]} vector, and Random is the average performance of the model after subtracting five random vectors from the \texttt{[CLS]} vector.}
%, while the model is robust to random perturbations in the last layer CLS token.
}
\label{tab:manipulation_roberta}
\end{table}

\section{Related work}
The explainability methods can be approached by local explanations and global explanations targeting post-hoc analysis or introducing interpretability in the architecture~\cite{madsen_post-hoc_2023, sundar2017, gradientInput, selvaraju_grad-cam_2020,kapishnikov_guided_2021, zhao_incorporating_2023,kim_interpretability_2018, ghorbani_towards_2019,jourdan_cockatiel_2023,zhao_explaining_2023,ribeiro-etal-2016-trust,rajagopal2021}. \citet{lyu_towards_2023} provides a %comprehensive 
survey of explainability methods in NLP. 
\LACOAT{} is a local explanation method providing post-hoc explanations given an input instance. 
Previous work attempted to explain and interpret NLP models using human-defined concepts~\cite{kim_interpretability_2018,CEBaB_neurips_2022} and concepts extracted from hidden representations~\cite{zhao_explaining_2023,ghorbani_towards_2019,rajani2020explaining,geva-etal-2022-transformer}. 
\citet{zhao_explaining_2023,kim_interpretability_2018} worked on the global explanation based on a surrogate model. %\citet{zhao_explaining_2023} trained the surrogate model using two optimization criteria i.e. auto encoding loss to stay faithful to the original model distribution and impact of the latent concept to prediction. 
%Different from them, w
We provide local explanations and we ensure the faithfulness of latent concepts 
%to the model 
by extracting them directly from the hidden representation without any supervised training. \citet{rajani2020explaining} used k-nearest neighbors of the training data %for low-confidence predictions and showed them to be useful 
to identify erroneous correlations and misclassified instances. 
%and enhancing the performance of the finetuned model. 
~\citet{dalvi2022discovering,sajjad-etal-2022-analyzing} analyzed latent concepts %of pretrained models 
in 
%terms of 
their ability to represent linguistic knowledge. 
Our \mone{} module is motivated by them. However, we propose a method to explain a model's prediction using %training data 
latent concepts.

\vspace{-1mm}
\section{Conclusion}
\vspace{-1mm}
We presented \LACOAT{} that provides a faithful and human-friendly explanation of a model's prediction. The qualitative evaluation and human evaluation showed that \LACOAT{} explanations are insightful in explaining a correct prediction, in highlighting a wrong prediction and in comparing the explanations of models. 
%
%using the training data latent space. 
The reliance on training data latent space enables interpreting how knowledge is structured in the network. Similarly, it enables the study of the evolution of predictions across layers.
%We performed a thorough qualitative and quantitative evaluation of each module of \LACOAT{}. 
%Moreover, \LACOAT{} enables the study of the evolution of knowledge 
%
%Moreover, \LACOAT{} has the potential to highlight potential biases in the 
\LACOAT{} promises human-in-the-loop in the decision-making process and is a step towards trust in AI. 
%understanding the prediction of the model

\section{Limitations}
We discussed the limitations of \LACOAT{} as follows: %1) 
%we did not establish a causal relationship between the selected latent concept and the prediction. In the paper, we concentrated on evaluating the concept behind \LACOAT{} and left the causal relation as future work. 2) 

\begin{itemize}
    \item While hierarchical clustering is better than nearest neighbor in discovering latent concepts as established by \citet{dalvi2022discovering}, it has computational limitations and it can not be easily extended to a corpus of say 1M tokens. However, the assumptions that are taken in the experimental setup e.g. considering the maximum 20 occurrences of a word (supported by \citet{dalvi2022discovering}) work well in practice in terms of limiting the number of tokens and covering all facets of a majority of the words. Moreover, the majority of the real-world tasks have limited task-specific data and \LACOAT{} can effectively be applied in such cases. In a contemporary work, \citet{hawasly-etal-2024-scaling} recently targeted this limitation by exploring alternative algorithms such as leaders and k-means, demonstrating that these are viable, cheaper alternatives to the expensive agglomerative hierarchical clustering.  Future explorations are needed to evaluate the correctness of the resulting clusters. For instance, k-means assume clusters to be spherical and similar assumption is enforced with the fixed distance threshold used in leaders. This limits their usefulness to identify clusters with irregular boundaries. 
    %could integrate these alternatives to enhance the Latent Concept Attribution method (LACOAT).

    \item \textcolor{black}{The concept discoverer is also constrained by a very small dataset. We conducted experiments to discover latent concepts with varying data sizes. The results indicate that when the data size is below 1000 instances, the discovered latent concepts tend to be sparse and lack coherence to form a clear concept.}

    \item \textcolor{black}{The computational cost of \LACOAT{} is higher than that of the IG method alone. While~\mone{} runs once per dataset as a pre-processing step, minimizing its impact on runtime during model inference, it still adds additional demand. Moreover,~\mthree{} is a linear classifier with an inference time of $O(d * c)$, where ${d}$ is the number of features and $c$ is the number of clusters. All other processes, aside from the IG part, have a linear cost.}

    \item  For tasks requiring reasoning over multiple sentences, we observe that sometimes the \LACOAT{} explanation's are not clearly indicative of the reason of a prediction which might be based on some syntactic and semantic similarity between multiple input sentences. A possible solution to this is to consider hierarchical relationship between latent concepts in contrast to considering a flat structure among latent concepts. The underlying setup of \mone{} supports this. However, comparing hierarchical structures requires further investigation beyond the scope of current work which provides a strong evidence towards faithful and human friendly explanations using training data latent space.
% may add something on playsifier module to have bias and cite relevant papers to use multiple models
    \item The human friendly explanations generated by \mfour{} are prone to errors due to the inherent limitations of generative models such as positional bias~\cite{NEURIPS2023_91f18a12, khan2024debating}, verbosity bias~\cite{huang2024limitations}, sensitivity to prompt style which hinders reproducibility. For instance, in this work, we observed that if we provide the predicted label to \mfour{}, it generates an explanation describing the relationship between input, latent concept and label irrespective of whether the label is correct or not. A possible solution is to use multiple LLMs~\cite{verga2024replacing,wang2023largelanguagemodelsfair,badshah2024referenceguidedverdictllmsasjudgesautomatic} and LLMs agents~\cite{chen2024autoagentsframeworkautomaticagent,tan2024peerreviewmultiturnlongcontext} that reasons over diverse explanations and reach to a better precise explanation of a latent concept. 

\end{itemize}

\section*{Acknowledgements}
We acknowledge the support of the Natural Sciences and Engineering Research Council of Canada (NSERC), RGPIN-2022-03943, Canada Foundation of Innovation (CFI) and Research Nova Scotia. Advanced computing resources are provided by ACENET, the regional partner in Atlantic Canada, and the Digital Research Alliance of Canada.  

\bibliography{main.bib}

\appendix

% \section{Appendix}
\label{sec:appendix}

\section{Datasets}
\label{app:datasets}
\begin{table}[!htbp]
\centering				
% \footnotesize
% \resizebox{\columnwidth}{!}{									
% \scalebox{1.0}{
% \setlength{\tabcolsep}{2.5pt}
    \begin{tabular}{l|c|c|c}					
    \toprule									
Task    & Train & Dev & Tags\\		
\midrule
    Sentiment & 13878 & 856 & 2\\
    POS & 36557 & 1802 & 48 \\
    Toxicity & 9000 & 800 & 2\\
    MNLI & 9000 & 1200 & 3\\
    \bottomrule
    \end{tabular}
    % }
% }
\caption{The data statistics of each dataset used in the evaluation experiments and the number of tags to be predicted. POS~\cite{marcus-etal-1993-building}, Jigsaw Toxicity dataset~\cite{cjadams_2017},  
the ERASER Sentiment dataset~\cite{eraser_sst,zaidan-eisner-2008-modeling} and the MNLI dataset~\cite{wang2019}}
\label{datasets_info}
\end{table}

\section{Finetuning Performance}
We tuned several transformers BERT-base-cased, RoBERTa and XLM-RoBERTa. We used standard splits for training, development and test data that we used to carry out our analysis. The splits to preprocess the data are available through git repository.\footnote{\url{https://github.com/nelson-liu/contextual-repr-analysis}} See Table~\ref{tab:pos_dataStats} and Table~\ref{tab:eraser_dataStats} for statistics and classifier accuracy. We present the results of Toxicity and MNLI in Appendix~\ref{app:toxicity} and~\ref{app:mnli}.

% \subsection{POS Sequence Tagger}
% \label{sec:appendix:tagger}
% We tuned several transformers BERT-base-cased, RoBERTa and XLM-RoBERTa. We used standard splits for training, development and test data that we used to carry out our analysis. The splits to preprocess the data are available through git repository\footnote{\url{https://github.com/nelson-liu/contextual-repr-analysis}}. See Table \ref{tab:pos_dataStats} for statistics and classifier accuracy. 

\begin{table}[h]
\centering				
% \footnotesize
\resizebox{\columnwidth}{!}{									
\scalebox{1.0}{
\setlength{\tabcolsep}{2.5pt}
    \begin{tabular}{l|cccc|c|c|c}									
    \toprule									
Task    & Train & Dev & Test & Tags & BERT & RoBERTa & XLM-R\\		
\midrule
    POS & 36557 & 1802 & 1963 & 48 & 96.81 & 96.70 & 96.75\\
    \bottomrule
    \end{tabular}
    }	
}
\caption{The fine-tuned performance of models, data statistics (number of sentences) on training, development, and test sets used in the finetuning, and the number of tags to be predicted for the POS tagging task. Model: BERT, RoBERTa, XLM-R}
\label{tab:pos_dataStats}
\end{table}

% \subsection{Sentiment Classification}
% \label{sec:appendix: sentiment}

\begin{table}[!htbp]
\centering				
% \footnotesize
\resizebox{\columnwidth}{!}{									
\scalebox{1.0}{
\setlength{\tabcolsep}{2.5pt}
    \begin{tabular}{l|cccc|c|c|c}									
    \toprule									
Task    & Train & Dev & Test & Tags & BERT & RoBERTa & XLM-R\\		
\midrule
    Sentiment & 13878 & 1516 & 2726 & 2 & 94.53 & 96.31 & 93.80\\
    \bottomrule
    \end{tabular}
    }
}
\caption{The fine-tuned performance of models, data statistics (number of sentences) on training, development, and test sets used in the finetuning, and the number of tags to be predicted for the sentiment classification task. Model: BERT, RoBERTa, XLM-R}
\label{tab:eraser_dataStats}		
\end{table}

\section{Qualitative Evaluation - More Examples}
\subsection{Example for the Evolution of Concepts} 
\begin{figure*}[]
\includegraphics[width=\linewidth]{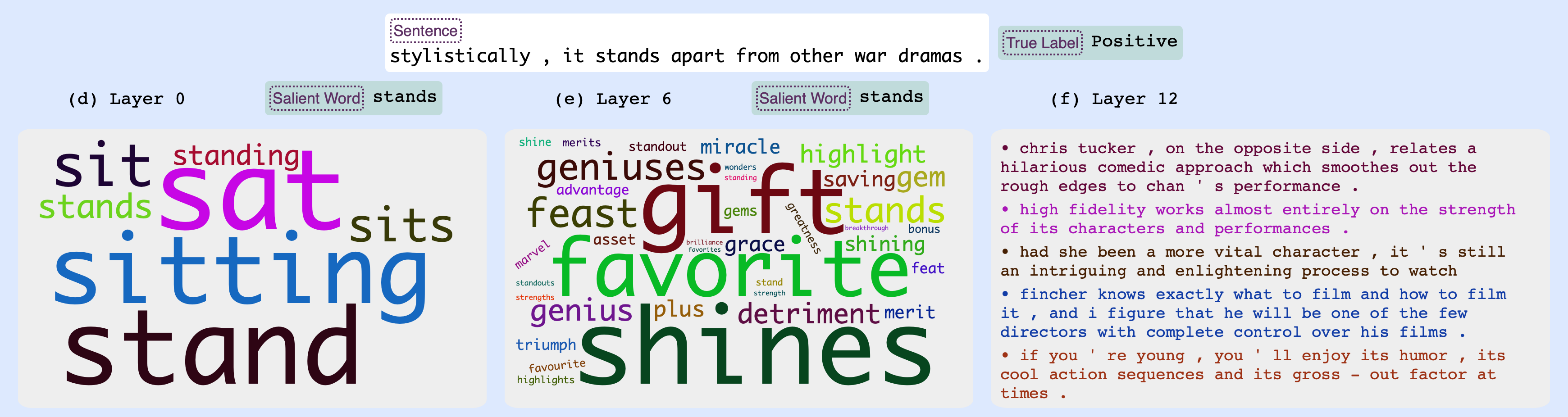}    
\caption{Sentiment task: Examples of the latent concepts of the most attributed words in layers 0, 6 and 12}
\label{fig:layersieexamples2}
\end{figure*}

Figure~\ref{fig:layersieexamples2} presents the other example of latent concepts of the salient words in layers 0, 6, and 12. Similarly to the example shown in Figure~\ref{fig:layersieexamples1}, the latent concept of this example shows that the different forms of the verb ``sit'' are not aligned with its usage in the input instance. The concept in the middle layer aligns better with the sentiment of the input sentence (Figure~\ref{fig:layersieexamples2}(b)). Most words of layer 6's latent concept match the sentiment of the input sentence. We also randomly pick five \texttt{[CLS]}  instances from the latent concept and show their corresponding sentences in the figure (see Figure~\ref{fig:layersieexamples2}(c)). The concept of the last layer is best aligned with the input sentence.

\subsection{Adversarial Example of the Sentence in Figure~\ref{subfig:eraser:b}}
\begin{figure}[!h]
\begin{center}
\includegraphics[width=0.9\columnwidth]{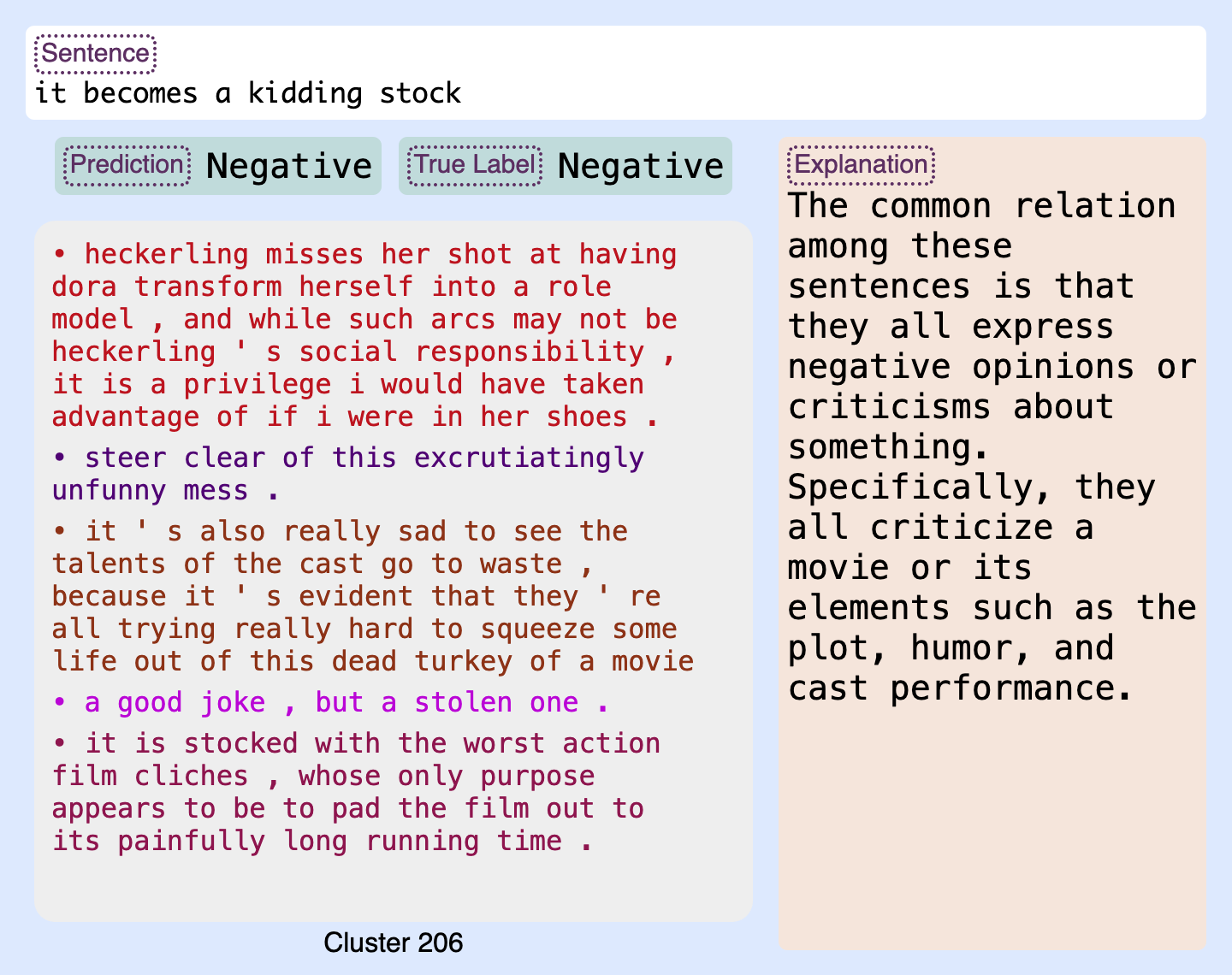}
\end{center}
\caption{An augmented example for the test instance in Figures~\ref{subfig:eraser:b}: The augmented sentence replaced the ``laughing'' with ``kidding'' which has a similar meaning. The label of the augmented sentence becomes positive, which is matched with the gold label. The new predicted latent concept is more closely aligned with the main sentence. The model may not learn the implicit meaning of the ``laughing stock'' in the sentence.}
\label{example4}
\end{figure}

The augmented sentence has a similar meaning word ``kidding'' instead of ``laughing'' (See Figure~\ref{example4}). The predicted label of the sentence becomes \texttt{Positive}, which is matched to the gold label. The latent concept of the ``kidding'' is more aligned with the sentence than the original one.

\subsection{Correct Predicted Label with Incorrect Gold Label}
\begin{figure}[]
\begin{center}
\includegraphics[width=0.9\columnwidth]{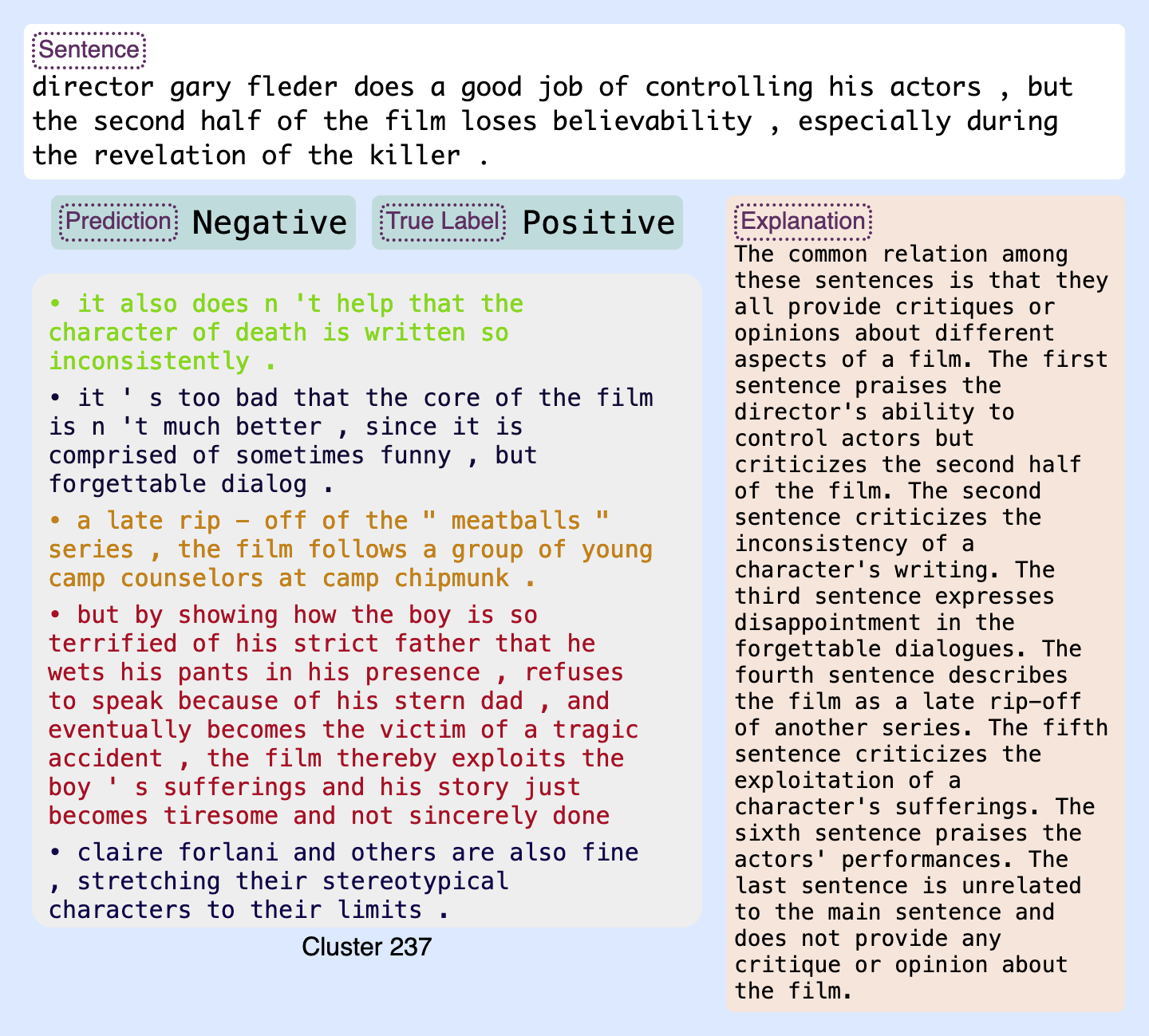}
\end{center}
\caption{A correct prediction but incorrect gold label: The test instance emphasizes the movie's shortcomings and uses the word "especially" to highlight the flaws. The explanation is rather long but it correctly highlights that the sentences are about ``critiques or opinions"}
\label{example5}
\end{figure}

The automatic labeling of latent concepts based on the model's class provides an opportunity to analyze the wrong predictions of the model with respect to the concept labels. We specifically observe the wrong predictions of test instances. We discovered that many of the wrong prediction cases were not caused by misclassification of the models but were due to the fact that the gold label was annotated incorrectly. Figure~\ref{example5} shows an example in which the main sentence and the explanation sentence share the same sentiment.  We can see that the sentence provides critiques of the different aspects of the film. But the gold label of this sentence is positive. We think the gold label for this sentence is incorrect.

%therefore while the model predicts a different label than the gold label, it is actually correct. Figure~\ref{example5} shows an example where the main sentence and the explanation sentence share the same sentiment. In addition, all sentences emphasize the film's shortcomings. According to the concept explanation, the given sentence would be considered a negative sentiment review. However, the gold label of this sentence is positive. Then, the sentence is checked by human understanding. We can see that the sentence provides critiques of the different aspects of the film. We think the gold label for this sentence is incorrect. This sentence was marked in a more accurate manner by the model than by the gold label.

\subsection{Incorrect Prediction in POS tagging Task}
\begin{figure}[!b]
\begin{center}
\includegraphics[width=0.9\columnwidth]{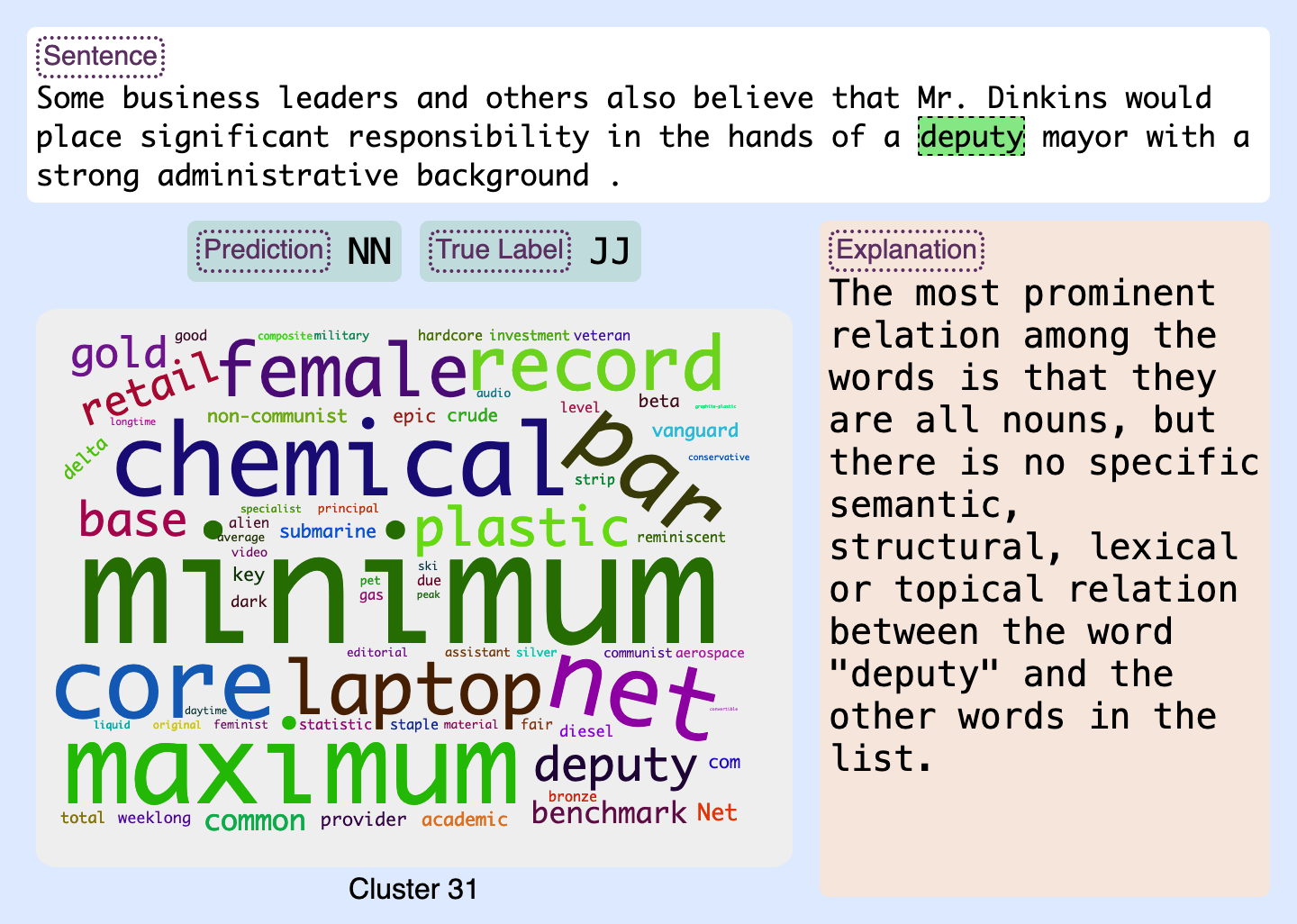}
\end{center}
\caption{An incorrect prediction (noun vs adjective) based on a latent concept made up of a mixture of nouns and adjectives: the ``deputy'' in this case is an adjective. The prediction aligns with a mixed cluster that contains both nouns and adjectives and the model may not learn to distinguish the ``noun'' and ``adjective'' in this case. The latent concept explanation is useful for the user to know that the model has used a mixed latent space for the prediction. The Explanation is rather wrong since it mentions that all these words are nouns. 
}
\label{posexample2}
\end{figure}

Figure~\ref{posexample2} presents an incorrect prediction in the POS tagging task. The prediction is aligned with a mixed concept that consists of nouns and adjectives. According to the latent concept explanation, we know that the model may not learn to distinguish the ``noun'' and ``adjective'', which causes the incorrect prediction. 

\begin{table*}[!t]
\centering
\footnotesize
\begin{tabular}{l|c|c|c|c|c|c|c|c|c} 
\hline
         & \multicolumn{9}{c}{\textbf{Sentiment}}                              \\ 
\hline
         & \multicolumn{3}{c|}{\textbf{BERT}}       &\multicolumn{3}{c|}{\textbf{RoBERTa}}    &\multicolumn{3}{c}{\textbf{XLM-R}}           \\ 
\hline
Layer    & \multicolumn{1}{l|}{Neg} & \multicolumn{1}{l|}{Pos} & \multicolumn{1}{l|}{Mix} & \multicolumn{1}{l|}{Neg} & \multicolumn{1}{l|}{Pos} & \multicolumn{1}{l|}{Mix} & \multicolumn{1}{l|}{Neg} & \multicolumn{1}{l|}{Pos} & \multicolumn{1}{l}{Mix}  \\ 
\hline
Layer 0  & 49 & 1  & 350 &45 & 0  & 355 &55 & 0  & 345                      \\
Layer 1  & 53 & 1  & 346 &50 & 0  & 350 &58 & 0  & 342                      \\
Layer 2  & 51 & 1  & 348 &49 & 0  & 351 &62 & 0  & 338                      \\
Layer 3  & 53 & 0  & 347 &60 & 0  & 340 &62 & 0  & 338                      \\
Layer 4  & 57 & 0  & 343 &52 & 0  & 348 &69 & 0  & 331                      \\
Layer 5  & 56 & 0  & 344 &51 & 0  & 349 &68 & 0  & 332                      \\
Layer 6  & 57 & 0  & 343 &45 & 1  & 354 &59 & 1  & 340                      \\
Layer 7  & 51 & 0  & 349 &56 & 2  & 342 &68 & 0  & 332                      \\
Layer 8  & 49 & 0  & 351 &116 &25 & 259 &71 & 0  & 329                      \\
Layer 9  & 66 & 4  & 330 &226 &126 &48  &82 & 7  & 311                      \\
Layer 10 & 125 &31 & 244 &235 &140 &25  &257  & 92 & 51                       \\
Layer 11 & 174  &49 & 177 &258 &120 &22  &256 & 110 &34                       \\
Layer 12 & 230  &81 & 89  &254  &126 &20  &105 & 270 &25                     \\
\hline
\end{tabular}
\caption{Number of clusters for each polarity: ``Neg'' for negative Label, ``Pos'' for positive, and ``Mix'' for mix label. The total number of clusters is 400.}
\label{app:tab:cluster_tag_numbers_eraser}
\end{table*}

\section{Module Specific Evaluation}
\subsection{\mone{} - Compositional Concept Examples}
\begin{figure}[!htbp]
    \begin{center}
    \begin{subfigure}[b]{0.47\linewidth}
\includegraphics[width=\linewidth]{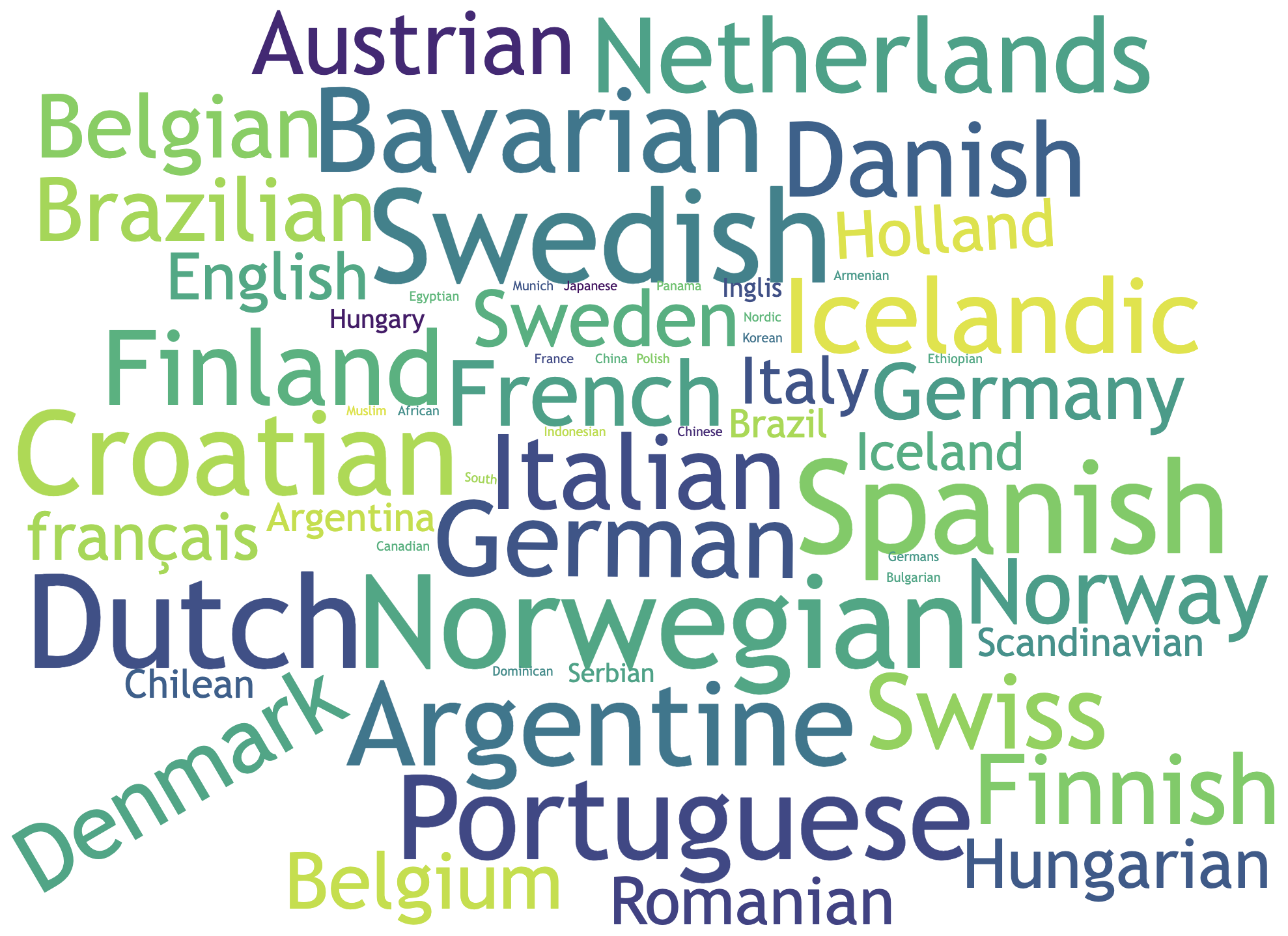}
    % \end{center}
    \caption{}
    \label{fig:jj+nn}
    \end{subfigure}
     \begin{subfigure}[b]{0.47\linewidth}
    % \begin{center}
    \includegraphics[width=\linewidth]{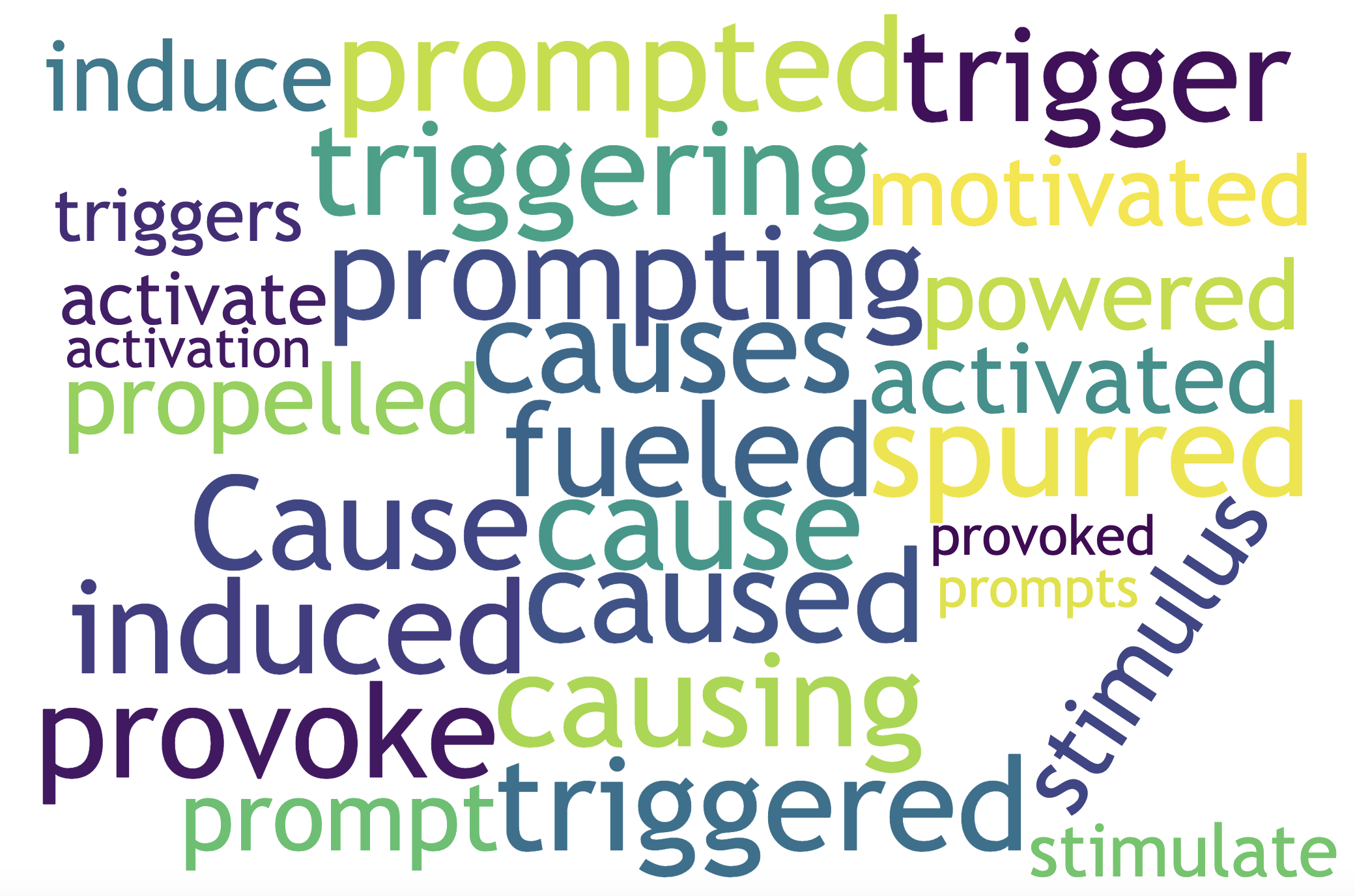}
    \caption{}
    \label{fig:verbForms}
    \end{subfigure}
    \end{center}
    \caption{Compositional concepts: (a) A cluster representing countries (NNP) and their adjectives (JJ), (b) Different form of verbs (Gerunds, Present and Past participles). 
    }
    \label{fig:clusters}
\end{figure}

We found that the concepts are not always formed aligning to the output class. Some concepts are formed by combining words from different classes. For example in Figure~\ref{fig:jj+nn}, the concept is composed of nouns (specifically countries) and adjectives that modify these country nouns. Similarly, Figure~\ref{fig:verbForms} describes a concept composed of different forms of verbs.
% \subsection{Compositional Concepts}

% We found that the concepts are not always formed aligning to the output class. Some concepts are formed by combining words from different classes. We demonstrate two examples below of concepts composed of different linguistic properties. For example in Figure \ref{fig:jj+nn}, the concept is composed of nouns (specifically countries) and adjectives that modify these country nouns. Similarly Figure \ref{fig:verbForms} describes a concept composed of different forms of verbs.

% Compositional Concepts are formed when 
% \hs{add info on compositional concept and discuss the concepts}

\subsection{\mone{} - Number of Clusters For Each Polarity in the Sentiment Classification Task}
Table~\ref{app:tab:cluster_tag_numbers_eraser} provides the number of clusters for each polarity in the sentiment classification task. It shows that the majority of latent concepts are class-based clusters at the last layer for the BERT, RoBERTa, and XLMR models.

\begin{table}[]
\begin{center}
\begin{tabular}{l|l|l|l}
    \toprule
    & \multicolumn{3}{c}{\textbf{POS}}  \\
    \midrule
    Layer & BERT & RoBERTa & XLM-R \\
    \midrule
    Layer 0	& 13.76	& 11.13 &	11.97 \\
Layer 1	& 12.75	& 13.58 &	11.91 \\
Layer 2	 & 15.51 & 15.60 & 12.99 \\
Layer 3	& 17.61	& 17.25 & 22.88 \\
Layer 4	& 23.81	& 20.30	& 32.08 \\
Layer 5 & 37.03 & 23.28 & 48.44 \\
Layer 6 & 64.83 & 32.52 & 67.94 \\
Layer 7 & 77.90	& 48.61 &	80.11 \\
Layer 8	& 86.96 & 73.88 &	85.83 \\
Layer 9	& 88.98 & 82.56 &	89.30 \\
Layer 10 &	89.99 &	83.24 &	89.94 \\
Layer 11 &	90.68 &	84.61 &	90.19 \\
Layer 12 &	92.16	& 85.67 &	90.18 \\
    \midrule
\end{tabular}
\caption{Saliency-based method (95\%): accuracy of \mtwo{} in mapping a representation to the correct latent concept in the POS tagging task. Model: BERT-base-cased, RoBERT-base, XLM-R}
\label{tab:pos_predicted_attribution_saliency}
\end{center}
\end{table}

% \begin{table}[]
% \centering
% \begin{tabular}{l|l|l|l}
%     \toprule
%     & \multicolumn{3}{c}{\textbf{Sentiment}}  \\
%     \midrule
%     Layer & BERT & RoBERTa & XLM-R \\
%     \midrule
% Layer 0 & 0	& 0 & 0 \\
% Layer 1	& 0	& 0 & 0 \\
% Layer 2	 & 0 & 0 & 0 \\
% Layer 3	& 0	& 0 & 0 \\
% Layer 4	& 0	& 0	& 0 \\
% Layer 5 & 0 & 0 & 0 \\
% Layer 6 & 0 & 0 & 0 \\
% Layer 7 & 0	& 0 & 0 \\
% Layer 8	& 0 & 99.11 & 0 \\
% Layer 9	& 37.09 & 98.45 & 0 \\
% Layer 10 & 99.55 & 99.14 & 0 \\
% Layer 11 & 99.82 & 99.27 & 99.17 \\
% Layer 12 & 99.25 & 99.27 & 99.08 \\
%     \midrule
% \end{tabular}
% \caption{Position-based method: accuracy of \mtwo{} in mapping a representation to the correct latent concept in the Sentiment Classification task. The reason of zero values is that the position-based method fails to find the right latent concept when the most attributed word is different from the position of the output head.}
% \label{tab:eraser_predicted_attribution_position}
% \end{table}

\begin{table}[]
\centering
\begin{tabular}{l|l|l|l}
    \toprule
    & \multicolumn{3}{c}{\textbf{Sentiment}}  \\
    \midrule
    Layer & BERT & RoBERTa & XLM-R \\
    \midrule
Layer 0 & 6.40	& 12.08 & 7.46 \\
Layer 1	& 7.12	& 12.46 & 5.57 \\
Layer 2	 & 7.66 & 17.29 & 6.36 \\
Layer 3	& 7.13	& 22.00 & 8.03 \\
Layer 4	& 12.18	& 20.08	& 9.71 \\
Layer 5 & 13.24 & 24.25 & 8.88 \\
Layer 6 & 11.18 & 17.26 & 8.75 \\
Layer 7 & 12.80	& 39.87 & 14.05 \\
Layer 8	& 4.06 & 92.84 & 15.75 \\
Layer 9	& 31.94 & 99.59 & 32.63 \\
Layer 10 &	99.57 &	99.69 &	92.06 \\
Layer 11 &	99.71 &	99.48 &	94.97 \\
Layer 12 &	99.25 & 99.27 &	99.08 \\
    \midrule
\end{tabular}
\caption{Saliency-based method: accuracy of \mtwo{} in mapping a representation to the correct latent concept in the sentiment classification task. The reason of very low values for the lower layers is mainly due to the absence of class-based latent concepts in the lower layers i.e. concepts that comprised more than 90\% of the tokens belonging to sentences of one of the classes.}
\label{tab:eraser_predicted_attribution_saliency}
\end{table}

\subsection{\mthree{} - Accuracy of \mthree{} for the Sentiment Classification and POS Tagging task}
We validate \mthree{} by measuring the accuracy of the test instances for both the sentiment classification and POS tagging tasks based on the BERT, RoBERTa, and XLMR models. The top 1, 2, and 5 accuracy of \mthree{} in mapping a representation to the correct latent concept for each layer is shown in Table~\ref{tab:pos_mapper_eval} and Table~\ref{tab:eraser_mapper_eval}. For all models, the performance of the top-5 is above 99\% for the POS tagging task and above 90\% for the sentiment classification task.

\onecolumn
\begin{table*}[]
\footnotesize
\label{tab:controlled_results_mapper}
\centering
\begin{tabular}{l|l|l|l|l|l|l|l|l|l}
    \toprule
        & \multicolumn{9}{c}{\textbf{POS}}  \\
        \midrule
        & \multicolumn{3}{c}{\textbf{BERT}} & \multicolumn{3}{c}{\textbf{RoBERTa}} & \multicolumn{3}{c}{\textbf{XLM-R}} \\
    \midrule
    Layer & Top-1 & Top-2 & Top-5 & Top-1 & Top-2 & Top-5 & Top-1 & Top-2 & Top-5 \\
    \midrule
    Layer 0 &  100	& 100	&  100 & 		99.91	& 99.95 &	99.98	&	99.99	& 100 & 100   \\
    Layer 1	& 100	& 100 &	100	 &	99.92 &	99.94 & 	99.98 &	100 &	100 &	100 \\
    Layer 2	& 100 &	100 &	100	& 99.76	& 99.92 &	99.98	&	99.72 &	99.98 &	100 \\
    Layer 3	& 99.85 &	99.98 &	100	&	99.38 &	99.85 &	99.98 &		98.25 &	99.60 &	99.98 \\
    Layer 4	& 99.72	& 99.92 &	99.97 &		98.67 &	99.58 &	99.87 &		97.72 &	99.60 &	99.98 \\
    Layer 5	& 99.03 &	99.75 &	99.94 &		97.69 &	99.15 &	99.73 &		97.05 &	99.23 &	99.91  \\
Layer 6	& 97.76	& 99.34 &	99.83 &	96.52 & 	98.71 &	99.59 &		95.8 &	98.95 &	99.76 \\
Layer 7	& 96.51	& 98.91 &	99.68 &	94.72 &	98.11 &	99.57 &		93.92 &	98.31 &	99.80 \\
Layer 8	& 95.27 &	98.52 &	99.79 &	92.56 &	97.55 &	99.52 &	94.20	& 98.52 &	99.80 \\
Layer 9	& 94.54 &	98.25 &	99.70 &	92.24 &	97.48 &	99.55 &	92.79 &	97.82 &	99.73  \\
Layer 10 & 92.67 & 97.89 &	99.68 &	91.61 &	97.19 &	99.55 &	92.03 & 97.66 &	99.60 \\
Layer 11 &	90.86 &	97.34 &	99.64 &	90.72 &	96.77 &	99.58 &	90.40 &	97.28 &	99.67  \\
Layer 12 & 84.19 & 94.15 &	99.05 &	86.88 &	95.13 &	99.15 &	85.07 &	94.57 &	99.08 \\
    \bottomrule
\end{tabular}
\caption{Top 1, 2, and 5 accuracy of ~\mthree{} in mapping a representation to the correct latent concept for the POS tagging task. The top-5 performance reaches above 99\% for all models demonstrating that the correct latent concept is among the top probable latent concepts of \mthree{}.}
\label{tab:pos_mapper_eval}
\end{table*}

\begin{table*}[]
\footnotesize
\centering
\begin{tabular}{l|l|l|l|l|l|l|l|l|l}
    \toprule
        & \multicolumn{9}{c}{\textbf{Sentiment}}  \\
        \midrule
        & \multicolumn{3}{c}{\textbf{BERT}} & \multicolumn{3}{c}{\textbf{RoBERTa}} & \multicolumn{3}{c}{\textbf{XLM-R}} \\
    \midrule
    Layer & Top-1 & Top-2 & Top-5 & Top-1 & Top-2 & Top-5 & Top-1 & Top-2 & Top-5 \\
    \midrule
0 & 100 & 100 & 100 & 99.95 & 100 & 100 & 100 & 100 & 100 \\
1 & 100 & 100 & 100 & 99.86 & 99.98 & 100 & 100 & 100 & 100 \\
2 & 100 & 100 & 100 & 99.89 & 99.98 & 100 & 99.9 & 100 & 100 \\
3 & 98.80 & 100 & 100 & 99.44 & 99.83 & 99.96 & 99.57 & 99.99 & 100\\
4 & 97.84 & 99.85 & 99.99 & 99.28 & 99.73 & 99.91 & 99.4 & 99.96 & 100 \\
5 & 97.19 & 99.63 & 99.94 & 98.4 & 99.5 & 99.84 & 99.12 & 99.84 & 99.96 \\
6 & 96.44 & 99.30 & 99.89 & 97.35 & 99.15 & 99.82 & 98.9 & 99.84 & 99.96 \\
7 & 94.86 & 98.97 & 99.90 & 96.13 & 98.74 & 99.63 & 98.22 & 99.62 & 99.9 \\
8 & 93.26 & 97.99 & 99.67 & 87.42 & 95.14 & 98.43 & 98.13 & 99.48 & 99.84 \\
9 & 90.42 & 96.97 & 99.20 & 75.38 & 88.14 & 96.07 & 96.37 & 98.77 & 99.66 \\
10 & 83.09 & 92.67 & 97.75 & 65.84 & 81.13 & 93.46 & 89.12 & 95.2 & 98.61 \\
11 & 76.84 & 88.02 & 96.01 & 65.91 & 81.36 & 93.43 & 70.99 & 84.31 & 94.18 \\
12 & 68.24 & 83.24 & 94.24 & 70.83 & 84.54 & 95.67 & 55.3 & 75.08 & 91.74 \\  \bottomrule
\end{tabular}
\caption{Top 1, 2, and 5 accuracy of ~\mthree{} in mapping a representation to the correct latent concept for the sentiment classification task. The top-5 performance reaches above 90\% for all models demonstrating that the correct latent concept is among the top probable latent concepts of \mthree{}.}
\label{tab:eraser_mapper_eval}
\end{table*}

\twocolumn
\section{Human Evaluation}
\label{app:humanevaluation}

\subsection{\LACOAT{} Effectiveness}
%To evaluate the explanation generated by \LACOAT{} empirically, 
We conduct a human evaluation using four annotators across 100 test samples. Specifically, given an explanation (e.g. Figure \ref{fig:eraser_pos_examples}), three annotators are asked to answer the following five questions: 

\begin{enumerate}[leftmargin=*,parsep=2pt,topsep=2pt]
    \item Regardless of the prediction, can you see any relation between the original input and the concept used by the model? (Yes/No)
    \vspace{-1mm}
    \item Given the prediction, does the \textit{latent concept} help you understand why the model made that prediction? (Helps/Neutral/Hinders)
    \vspace{-1mm}
    \item Given the prediction, does the \textit{explanation} help you understand why the model made that prediction? (Helps/Neutral/Hinders)
    \vspace{-1mm}
    \item Does the explanation \textit{accurately} describe the latent concept? (Yes/No)
    \vspace{-1mm}
    \item Is the explanation \textit{relevant} to the task at hand? (Yes/No)
    
\end{enumerate}

\subsection{Comparison with other Methods}
For comparison with other methods, we ask four annotators to rank 100 samples where they see the original input, gold label, predicted label, and explanations by three methods: LACOAT, IG and COCKATIEL. \textcolor{black}{We consider the optimal settings for each explanation method. IG explanations are shown for layer 0 based on salient input features, while in layer 12, IG identifies only the [CLS] token as the most salient feature, which provides an implausible explanation and may negatively impact its human evaluation. COCKATIEL explanations are restricted to layer 12. LACOAT explanations span three layers (0, 6, and 12). By presenting LACOAT's layerwise explanations, we offer a more comprehensive view of how the prediction evolves through the network, enhancing human evaluation.}

The annotators are asked to rank each method from 1 to 3 in terms of usefulness in understanding the reason for the prediction where 1 implies the method was very useful while 3 implies it was not useful. The annotation allows for the annotator to rank multiple methods with the same usefulness rating, e.g. for a particular sample, both LACOAT and COCKATIEL can have the rank 1. This setting is intentional since the output of explanation methods is not directly comparable to each other due to the difference in their design and the targeted form and granularity of explanation. Table~\ref{tab:human-eval} presents the results. The results suggested that \LACOAT{} is preferred or equally preferred by all annotators. The average Cohen's $\kappa$ further shows a "fair agreement" between annotators and the consolidated ranking where consolidated ranking is the average rank across users.

\section{Toxicity Classification Task}
\label{app:toxicity}
\subsection{Experimental Setup}
We use the Jigsaw Toxicity dataset for the toxicity classification task (Toxicity). This dataset comprises Wikipedia comments labeled by human annotators to identify instances of toxic behavior. We retain only the "toxic" feature as the label for each instance, thereby classifying each instance as \texttt{toxic} or \texttt{non-toxic}. The dataset has more than 159k, 63k, and 89k instances for train, dev, and test. We randomly select 9k, 800, and 800 splits for train, dev, and test respectively. We use $K=600$ for \mone{} and have the same setting for the rest of the module-specific hyperparameters. 

We also used standard splits to tune transformers BERT-base-cased, RoBERTa, and XLM-RoBERTa. The fine-tuned performance of each model is presented in Table~\ref{tab:toxicity_dataStats}.

\begin{table}[!htbp]
\centering				
% \footnotesize
\resizebox{\columnwidth}{!}{									
\scalebox{1.0}{
\setlength{\tabcolsep}{2.5pt}
    \begin{tabular}{l|cccc|c|c|c}									
    \toprule									
Task    & Train & Dev & Test & Tags & BERT & RoBERTa & XLM-R\\		
\midrule
    Toxicity & 159570 & 63977 & 89185 & 2 & 91.53 & 91.55 & 91.53\\
    \bottomrule
    \end{tabular}
    }
}
\caption{The fine-tuned performance of models, data statistics (number of sentences) on training, development, and test sets used in the finetuning, and the number of tags to be predicted for the toxicity classification task. Model: BERT, RoBERTa, XLM-R}
\label{tab:toxicity_dataStats}		
\end{table}

\subsection{Qualitative Evaluation}
\subsubsection{Correct prediction with correct gold label}
\begin{figure}[!h]
\begin{center}
\includegraphics[width=\linewidth]{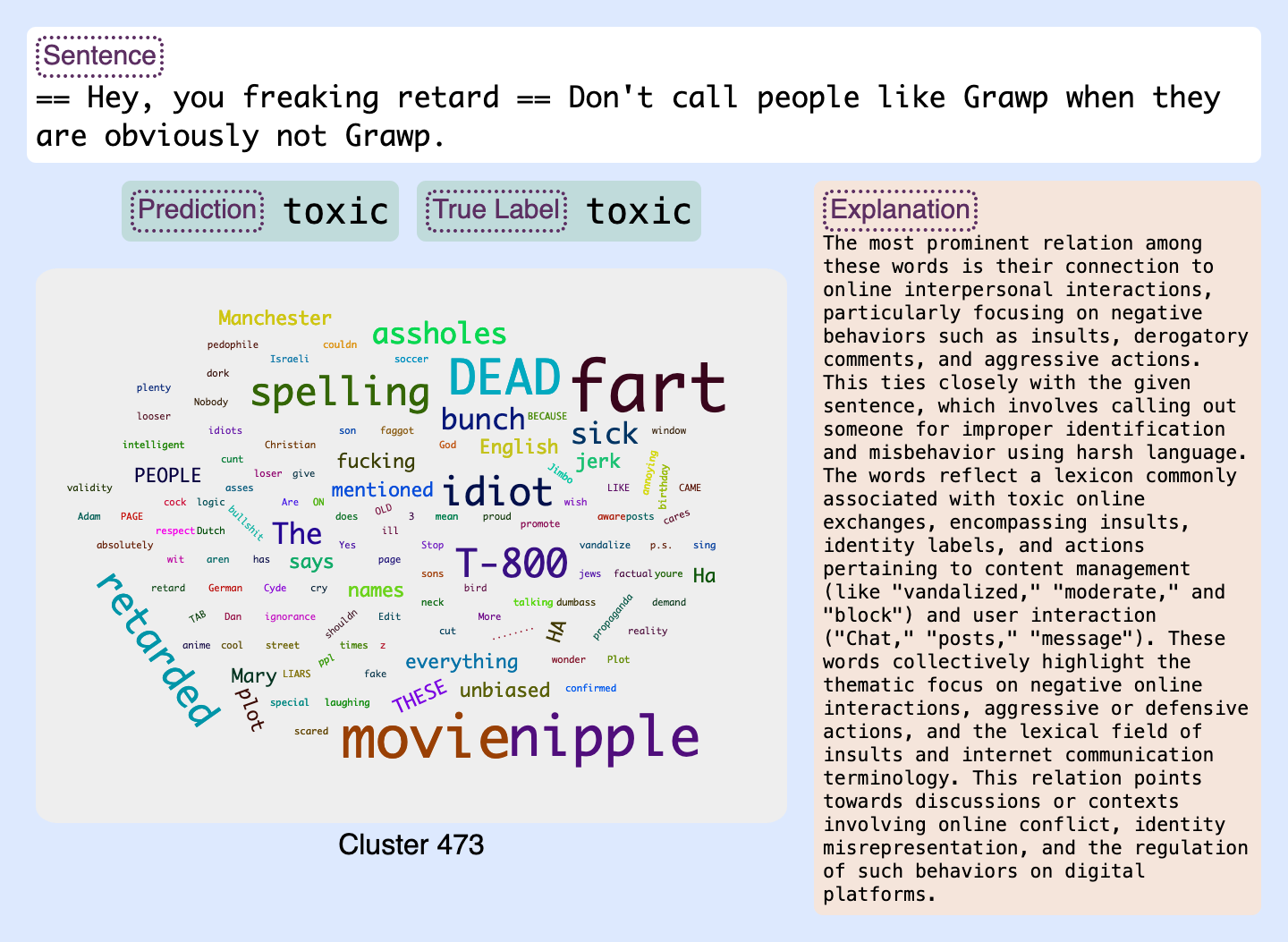}
\end{center}
\caption{RoBERTa: A toxic labeled test instance correctly predicted by the model.}
\label{app:toxicity_correct_prediction_example}
\end{figure}

\begin{figure}[!h]
\begin{center}
\includegraphics[width=\linewidth]{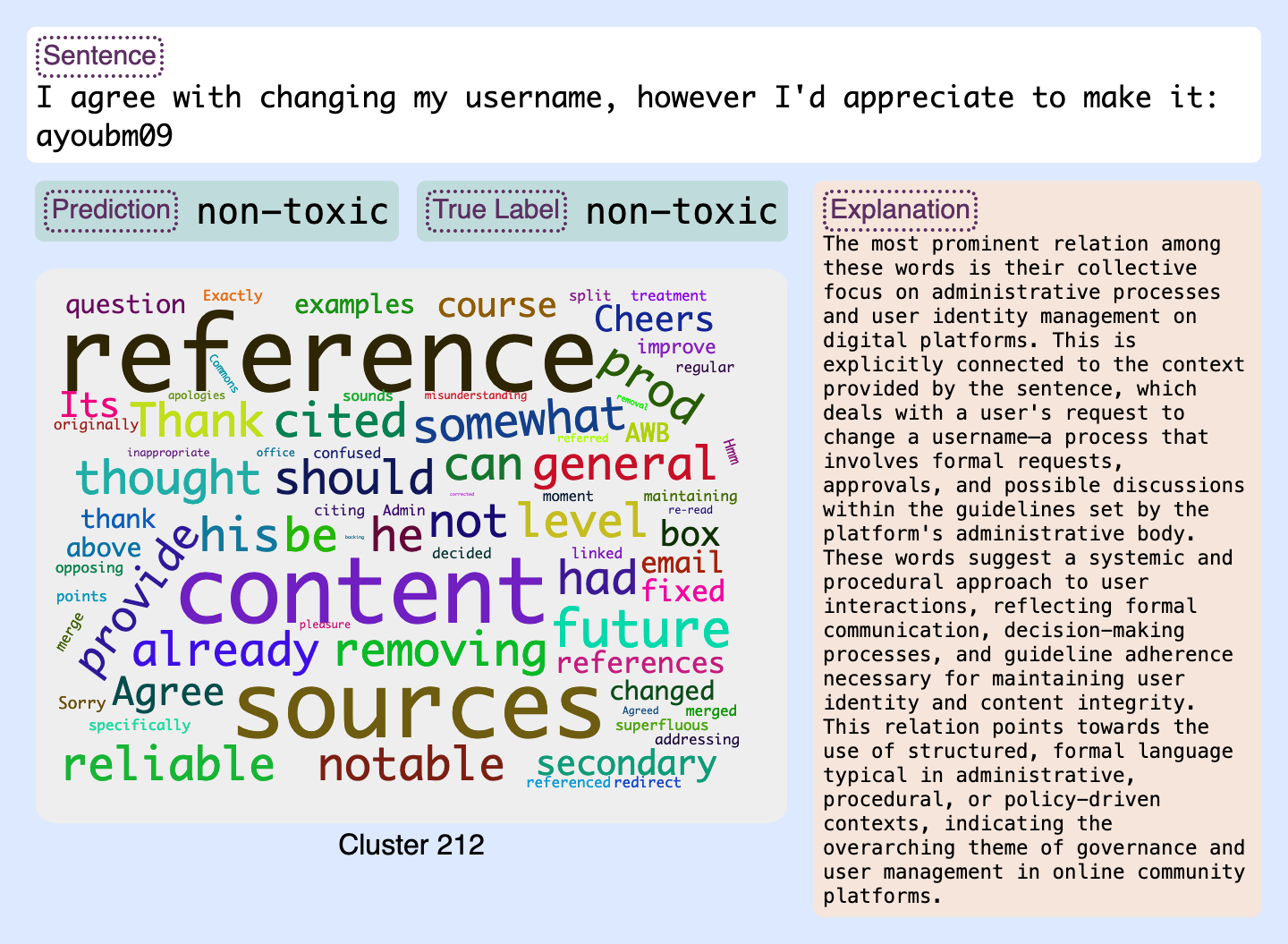}
\end{center}
\caption{RoBERTa: A non-toxic labeled test instance correctly predicted by the model.}
\label{app:toxicity_correct_nontoxic_example}
\end{figure}

Figure~\ref{app:toxicity_correct_prediction_example} and Figure~\ref{app:toxicity_correct_nontoxic_example} present the correct prediction case for a toxic and a non-toxic labeled instance. In the toxic label instance, \mfour discovers that the words in latent concept have common semantics of negative behaviors and highlights the reason for toxic label due to harsh language. For the non-toxic labeled instance, \mfour finds that the relation between the sentence and the list of words in the latent concept is about the governance theme and user management in online community platforms.

\subsubsection{Wrong prediction with correct gold label}
\begin{figure}[!h]
\begin{center}
\includegraphics[width=\linewidth]{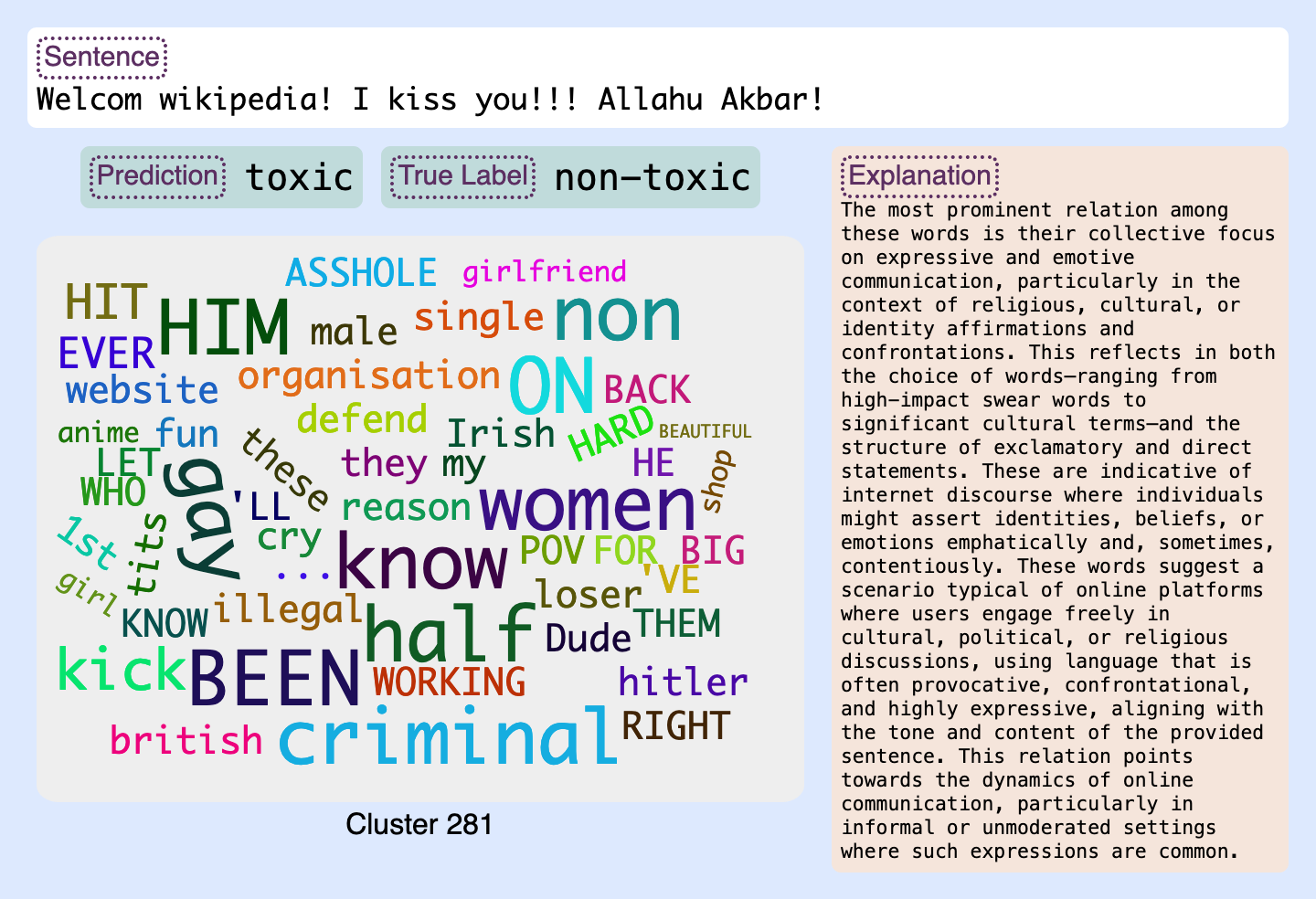}
\end{center}
\caption{RoBERTa: A non-toxic labeled instance that is incorrectly predicted as toxic.}
\label{app:toxicity_incorrect_example}
\end{figure}

Figure~\ref{app:toxicity_incorrect_example} shows a non-toxic labeled instance that is incorrectly predicted as toxic. The sentence contains non-toxic content and has cultural/religious terms expressing positive emotion. However, the model predicts this sentence with a toxic label. The latent concept provides helpful evidence that it contains many toxic words such as ``ASSHOLE'', ``idiot'', ``bitch'', and ``Niggers''. Also, the \mfour provides additional information that both the sentence and the latent concept contain the context of religion and culture. We hypothesize that the model captures the correlations between the toxic content or label and the religion/culture concept in the training. Thus, the model has a bias in the prediction with the religion/culture-related content to the toxic label.

% \subsubsection{Correct Predicted Label with Incorrect Gold Label}
% We find many wrong prediction instances caused by incorrect gold labels in this task. For example, Figure~\ref{app:toxicity_incorrect_annotation_example} presents an incorrect annotated instance. \mfour detects that the main sentence and a couple of sentences in latent concepts are related to the request for intervention or assistance and points out that the main sentence is a general help request. In other words, the main sentence should be labeled as non-toxic.

\subsection{Module Specific Evaluation}
\subsubsection{\mone{}}
We also form latent concepts of each layer using \mone{} and annotate them with the procedure mentioned in \ref{text:annotation}. In the toxicity classification task, we discovered that 88\%, 99\%, and 96\% of the latent concepts of BERT, RoBERTa, and XLMR were made up of either toxic majority or non-toxic majority sentences (see Table~\ref{app:tab:cluster_tag_numbers_toxicity}). Similar to the sentiment, we noticed that the 12th layer has a higher number of class-based clusters of Roberta and XLMR.

\subsubsection{\mtwo{}}
For toxicity, we found over 98\% accuracy in mapping the salient representation to the correct latent concept for the last layer (see Tables~\ref{tab:toxicity_predicted_attribution_saliency}). This high accuracy indicates that \mtwo{} performs effectively and accurately in the toxicity task.

\subsubsection{\mthree{}}
Table~\ref{tab:toxicity_mapper_eval} presents the performance of \mthree{} for toxicity. The accuracy of the first layer is high (around 100\%) and drops as the layer increases for all models. In the last layer, the accuracy of the top prediction arrives at 67.01\%, 81.43\%, and 64.19\% for BERT, RoBERTa, and XLMR. We also consider the top two and top five predictions of the mapper. The performances of the top two and the top five predictions are more than 81\% and 93\% for these three models. Especially, the mapper based on the RoBERTa model has the best performance, achieving 81.43\%, 93.72\%, and 98.21\% for the top one, two, and five predictions respectively.

% \begin{table}[]
% \centering
% \begin{tabular}{l|l|l|l}
%     \toprule
%     & \multicolumn{3}{c}{\textbf{Toxicity}}  \\
%     \midrule
%     Layer & BERT & RoBERTa & XLM-R \\
%     \midrule
% Layer 0 & 0	& 0 & 0 \\
% Layer 1	& 0	& 0 & 0 \\
% Layer 2	 & 0 & 0 & 0 \\
% Layer 3	& 46.90	& 0 & 0 \\
% Layer 4	& 47.63	& 0	& 0 \\
% Layer 5 & 19.93 & 0 & 0 \\
% Layer 6 & 61.82 & 0 & 0 \\
% Layer 7 & 66.43	& 94.98 & 0 \\
% Layer 8	& 64.00 & 97.37 & 78.13 \\
% Layer 9	& 80.14 & 97.62 & 98.72 \\
% Layer 10 & 92.99 & 98.80 & 96.59 \\
% Layer 11 & 94.83 & 99.47 & 96.65 \\
% Layer 12 & 98.93 & 99.72 & 99.61 \\
%     \midrule
% \end{tabular}
% \caption{Position-based method: accuracy of \mtwo{} in mapping a representation to the correct latent concept in the Toxicity Classification task. The reason of zero values is that the position-based method fails to find the right latent concept when the most attributed word is different from the position of the output head.}
% \label{tab:toxicity_predicted_attribution_position}
% \end{table}

\onecolumn
\begin{table*}[!t]
\centering
\footnotesize
\begin{tabular}{l|c|c|c|c|c|c|c|c|c} 
\hline
         & \multicolumn{9}{c}{\textbf{Toxicity}}                              \\ 
\hline
         & \multicolumn{3}{c|}{\textbf{BERT}}       &\multicolumn{3}{c|}{\textbf{RoBERTa}}    &\multicolumn{3}{c}{\textbf{XLM-R}}           \\ 
\hline
Layer    & \multicolumn{1}{l|}{non-toxic} & \multicolumn{1}{l|}{toxic} & \multicolumn{1}{l|}{Mix} & \multicolumn{1}{l|}{non-toxic} & \multicolumn{1}{l|}{toxic} & \multicolumn{1}{l|}{Mix} & \multicolumn{1}{l|}{non-toxic} & \multicolumn{1}{l|}{toxic} & \multicolumn{1}{l}{Mix}  \\ 
\hline
Layer 0  & 15 & 30  & 555 &22 & 15  & 563 & 19 & 16  & 565                      \\
Layer 1  & 13 & 27  & 560 &17 & 20  & 563 & 16 & 16  & 568                      \\
Layer 2  & 11 & 33  & 556 &18 & 24  & 558 & 16 & 20  & 564                      \\
Layer 3  & 16 & 35  & 549 &17 & 28  & 555 & 16 & 21  & 563                      \\
Layer 4  & 18 & 36  & 546 &20 & 29  & 551 & 15 & 24  & 561                      \\
Layer 5  & 12 & 41  & 547 &28 & 33  & 539 & 14 & 22  & 564                      \\
Layer 6  & 15 & 48  & 537 &37 & 42  & 521 & 23 & 24  & 553                      \\
Layer 7  & 18 & 49  & 533 &324 & 131  & 145 & 114 & 53  & 433                      \\
Layer 8  & 23 & 49  & 528 &332 & 186 & 82 & 267 & 74 & 259                      \\
Layer 9  & 43 & 52  & 505 &373 & 158 &69  & 334 & 134  & 132                      \\
Layer 10 & 116 & 73 & 411 &425 &137 &38  & 328  & 154 & 118                       \\
Layer 11 & 298  & 110 & 192 &449 &130 &21  &423 & 139 & 38                       \\
Layer 12 & 374  & 155 & 71  &502  &92 &6  &449 & 129 & 22                    \\
\hline
\end{tabular}
\caption{Number of clusters for each polarity. The total number of clusters is 600.}
\label{app:tab:cluster_tag_numbers_toxicity}
\end{table*}

\begin{table*}[]
\footnotesize
\centering
\begin{tabular}{l|l|l|l|l|l|l|l|l|l}
    \toprule
        & \multicolumn{9}{c}{\textbf{Toxicity}}  \\
        \midrule
        & \multicolumn{3}{c}{\textbf{BERT}} & \multicolumn{3}{c}{\textbf{RoBERTa}} & \multicolumn{3}{c}{\textbf{XLM-R}} \\
    \midrule
    Layer & Top-1 & Top-2 & Top-5 & Top-1 & Top-2 & Top-5 & Top-1 & Top-2 & Top-5 \\
    \midrule
0 & 100 & 100 & 100 & 99.96 & 99.99 & 100 & 100 & 100 & 100 \\
1 & 100 & 100 & 100 & 99.92 & 100 & 100 & 100 & 100 & 100 \\
2 & 99.99 & 100 & 100 & 99.94 & 100 & 100 & 99.75 & 100 & 100 \\
3 & 99.07 & 99.88 & 100 & 99.34 & 99.80 & 99.92 & 99.46 & 99.95 & 100\\
4 & 98.49 & 99.78 & 99.99 & 96.87 & 98.96 & 99.78 & 98.81 & 99.83 & 100 \\
5 & 98.25 & 99.72 & 99.94 & 93.10 & 97.63 & 99.26 & 97.72 & 99.42 & 99.89 \\
6 & 97.22 & 99.51 & 99.88 & 87.72 & 95.05 & 98.50 & 94.83 & 98.45 & 99.61 \\
7 & 95.00 & 98.57 & 99.68 & 73.50 & 87.21 & 95.70 & 86.96 & 95.37 & 98.72 \\
8 & 91.87 & 97.41 & 99.18 & 67.62 & 83.09 & 94.38 & 79.62 & 91.37 & 97.62 \\
9 & 85.66 & 93.80 & 98.01 & 66.75 & 82.80 & 94.38 & 73.73 & 88.57 & 96.76 \\
10 & 76.22 & 87.90 & 95.89 & 64.87 & 81.37 & 93.07 & 66.10 & 82.36 & 93.39 \\
11 & 70.53 & 84.31 & 94.31 & 77.91 & 91.09 & 98.10 & 68.30 & 84.49 & 95.28 \\
12 & 67.01 & 81.71 & 93.65 & 81.43 & 93.72 & 98.21 & 64.19 & 81.96 & 94.26 \\  \bottomrule
\end{tabular}
\caption{Top 1, 2, and 5 accuracy of ~\mthree{} in mapping a representation to the correct latent concept for the toxicity classification task. The top-5 performance reaches above 90\% for all models demonstrating that the correct latent concept is among the top probable latent concepts of \mthree{}.}
\label{tab:toxicity_mapper_eval}
\end{table*}

\twocolumn
\begin{table}[]
\centering
\footnotesize
\begin{tabular}{l|l|l|l}
    \toprule
    & \multicolumn{3}{c}{\textbf{Toxicity}}  \\
    \midrule
    Layer & BERT & RoBERTa & XLM-R \\
    \midrule
Layer 0 & 10.54	& 13.45 & 6.57 \\
Layer 1	& 8.98	& 19.14 & 8.45 \\
Layer 2	 & 10.92 & 19.92 & 10.56 \\
Layer 3	& 49.90	& 22.95 & 13.90 \\
Layer 4	& 50.07	& 34.30	& 15.12 \\
Layer 5 & 11.30 & 31.50 & 23.89 \\
Layer 6 & 66.21 & 35.42 & 34.47 \\
Layer 7 & 67.11	& 91.84 & 59.38 \\
Layer 8	& 63.74 & 97.84 & 77.43 \\
Layer 9	& 84.41 & 98.79 & 94.44 \\
Layer 10 &	94.92 &	99.30 &	97.52 \\
Layer 11 &	94.73 &	99.49 &	97.39 \\
Layer 12 &	98.93 & 99.72 &	99.61\\
    \midrule
\end{tabular}
\caption{Saliency-based method: accuracy of \mtwo{} in mapping a representation to the correct latent concept in the toxicity classification task. The reason of very low values for the lower layers is mainly due to the absence of class-based latent concepts in the lower layers i.e. concepts that comprised more than 90\% of the tokens belonging to sentences of one of the classes.}
\label{tab:toxicity_predicted_attribution_saliency}
\end{table}

\begin{figure}[]
\begin{center}
\includegraphics[width=\linewidth]{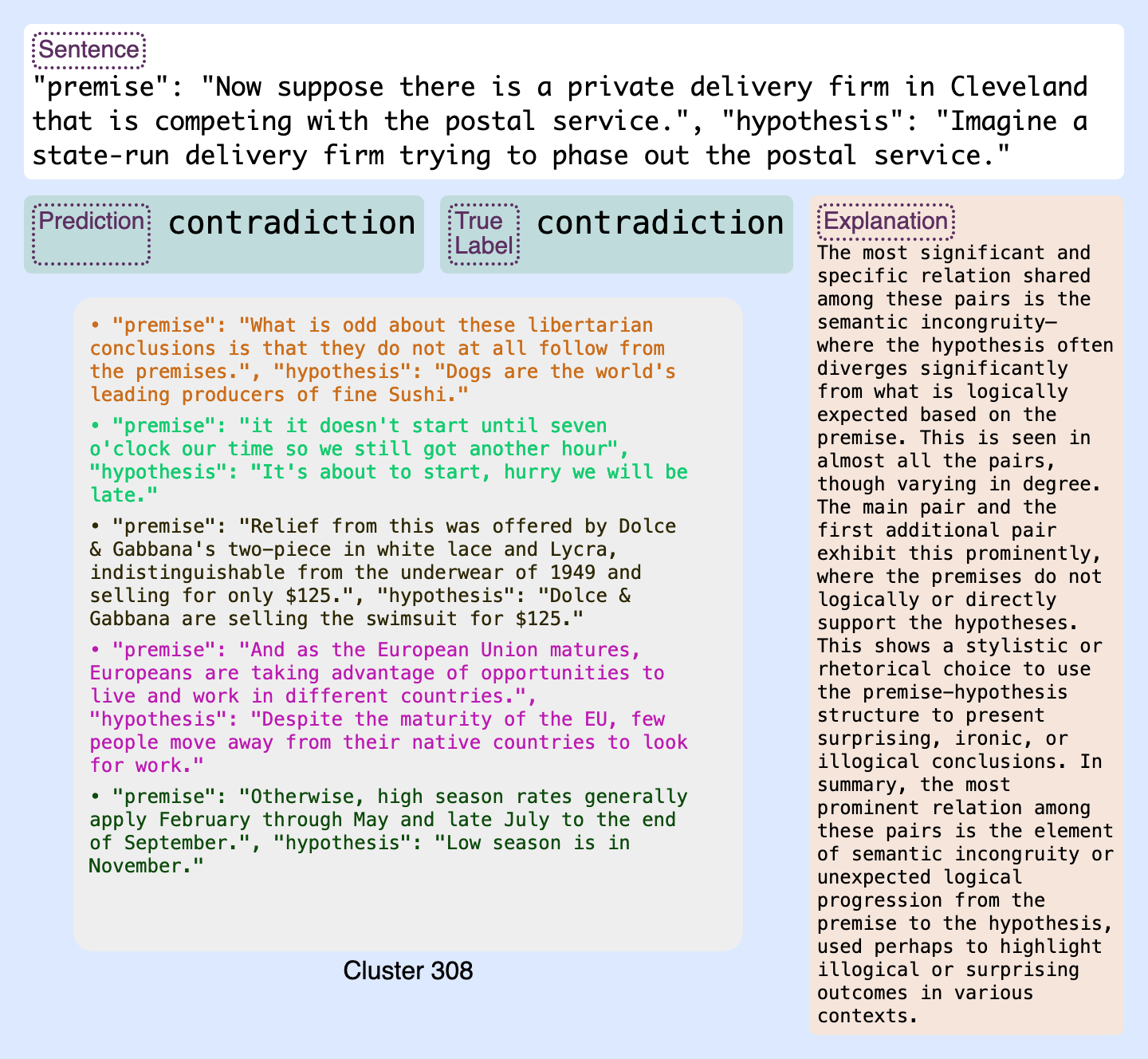}
\end{center}
\caption{MNLI: A contradiction labeled instance that is correctly predicted.}
\label{app:mnli_correct_example}
\end{figure}

%\onecolumn

% \begin{figure}[!h]
% \begin{center}
% \includegraphics[width=\linewidth]{figures/toxicity/correct_prediction_incorrect_annotation.png}
% \includegraphics[width=\linewidth]{figures/toxicity/plausifyer_incorrect_annotation.png}
% \end{center}
% \caption{RoBERTa: A non-toxic labeled instance that is incorrectly predicted as toxic.}
% \label{app:toxicity_incorrect_annotation_example}
% \end{figure}

% \begin{figure}[!h]
% \begin{center}
% \includegraphics[width=\linewidth]{figures/toxicity/correct_prediction_incorrect_annotation.png}
% \includegraphics[width=\linewidth]{figures/toxicity/plausifyer_incorrect_annotation.png}
% \end{center}
% \caption{RoBERTa: A non-toxic labeled instance that is incorrectly predicted as toxic.}
% \label{app:toxicity_incorrect_annotation_example}
% \end{figure}

%\twocolumn
\section{NLI Task}
\label{app:mnli}
\subsection{Experimental Setup}
We use the MNLI dataset for the NLI task. This task classifies each sentence pair into three classes: \texttt{entailment}, \texttt{contradiction}, and \texttt{neutral}. The MNLI dataset contains 393k, 19.65k, and 19.65k splits for train, dev, and test. We randomly select 9k and 1.2k for train and dev splits. We use $K = 400$ for \mone{} and set the same numbers for the other hyperparameters. 

Like the other task, we used standard splits to tune transformers BERT-base-cased, RoBERTa, and XLM-RoBERTa. The fine-tuned performance of each model is presented in Table~\ref{tab:mnli_dataStats}.

\begin{table}[!htbp]
\centering				
% \footnotesize
\resizebox{\columnwidth}{!}{									
\scalebox{1.0}{
\setlength{\tabcolsep}{2.5pt}
    \begin{tabular}{l|cccc|c|c|c}									
    \toprule									
Task    & Train & Dev & Test & Tags & BERT & RoBERTa & XLM-R\\		
\midrule
    MNLI & 393000 & 19650 & 19650 & 3 & 84.00 & 87.69 & 84.54 \\
    \bottomrule
    \end{tabular}
    }
}
\caption{The fine-tuned performance of models, data statistics (number of sentences) on training, development, and test sets used in the finetunings, and the number of tags to be predicted for the MNLI task. Model: BERT, RoBERTa, XLM-R}
\label{tab:mnli_dataStats}		
\end{table}

\subsection{Qualitative Evaluation}
Figure~\ref{app:mnli_correct_example} shows a correct prediction instance with a ``contradiction'' label. \mfour{} detects that all premise-hypothesis pairs are ``semantic incongruity'', which means that the premise sentence does not have a matched logic with the hypothesis sentence. This indicates that the model learns the knowledge of the ``contradiction" label in the training.

However, due to the complexity of the task, it is difficult for humans to understand or find the relationship between the latent concept and the prediction of the input sentence. Especially, if we have the word cloud as the latent concept-based explanation, it may not be helpful for humans to interpret the model prediction. \mfour{} simplifies the interpretation in such cases.

\subsection{Module Specific Evaluation}
\subsubsection{\mone{}}
In the MNLI task, we found more ``mixed'' latent concepts than class-based latent concepts related to other tasks. There are 0\%, 82\%, and 58\% discovered label dominant latent concepts by BERT, RoBERTa, and XLMR (see Table~\ref{app:tab:cluster_tag_numbers_mnli}). We speculate that tasks that involve multiple sentences as input are more complex and abstract, thereby it is difficult to have clear distinct concepts. This observation also varies depending on the model. For instance, we did not detect any class-based latent concepts of the BERT model. However, we achieve good performance in discovering the latent concept when using the RoBERTa model.

% \twocolumn
\subsubsection{\mtwo{}}
We found that both RoBERTa and XLMR models have over 90\% accuracy for the salient representation mapping for the last layer (see Tables~\ref{tab:mnli_predicted_attribution_saliency}). To some extent, this accuracy indicates that \mtwo{} have good performance in the MNLI task based on the RoBERTa and XLMR model. Unlike other tasks, we have extremely low accuracy with the BERT model. We assume that the BERT model may not be able to capture the task knowledge due to the task complexity. 

% \begin{table}[]
% \centering
% \begin{tabular}{l|l|l|l}
%     \toprule
%     & \multicolumn{3}{c}{\textbf{MNLI}}  \\
%     \midrule
%     Layer & BERT & RoBERTa & XLM-R \\
%     \midrule
% Layer 0 & 0	& 0 & 0 \\
% Layer 1	& 0	& 0 & 0 \\
% Layer 2	 & 0 & 0 & 0 \\
% Layer 3	& 0	& 0 & 0 \\
% Layer 4	& 0	& 0	& 0 \\
% Layer 5 & 0 & 0 & 0 \\
% Layer 6 & 0 & 0 & 0 \\
% Layer 7 & 0	& 0 & 0 \\
% Layer 8	& 0 & 0 & 0 \\
% Layer 9	& 0 & 30.54 & 0 \\
% Layer 10 & 0 & 85.29 & 0 \\
% Layer 11 & 0 & 87.5 & 81.38 \\
% Layer 12 & 0 & 95.22 & 90.58 \\
%     \midrule
% \end{tabular}
% \caption{Position-based method: accuracy of \mtwo{} in mapping a representation to the correct latent concept in the MNLI task. The reason of zero values is that the position-based method fails to find the right latent concept when the most attributed word is different from the position of the output head.}
% \label{tab:mnli_predicted_attribution_position}
% \end{table}

\begin{table}[]
\centering
\begin{tabular}{l|l|l|l}
    \toprule
    & \multicolumn{3}{c}{\textbf{MNLI}}  \\
    \midrule
    Layer & BERT & RoBERTa & XLM-R \\
    \midrule
Layer 0 & 0.027	& 0.41 & 0.56 \\
Layer 1	& 0.083	& 0.67 & 0.43 \\
Layer 2	 & 0.04 & 0 & 0.23 \\
Layer 3	& 0	& 0.05 & 0.35 \\
Layer 4	& 0.10	& 0	& 0.08 \\
Layer 5 & 0.10 & 0 & 0.12 \\
Layer 6 & 0.05 & 0 & 0.12 \\
Layer 7 & 0	& 0 & 0.13 \\
Layer 8	& 0 & 21.61 & 0 \\
Layer 9	& 0 & 83.90 & 14.29 \\
Layer 10 &	0 &	91.78 &	55.93 \\
Layer 11 &	0 &	92.63 &	89.73 \\
Layer 12 &	0 & 95.22 &	90.58\\
    \midrule
\end{tabular}
\caption{Saliency-based method: accuracy of \mtwo{} in mapping a representation to the correct latent concept in the MNLI task. The reason of very low values for the lower layers is mainly due to the absence of class-based latent concepts in the lower layers i.e. concepts that comprised more than 90\% of the tokens belonging to sentences of one of the classes.}
\label{tab:mnli_predicted_attribution_saliency}
\end{table}

\subsubsection{\mthree{}}
Similar to other tasks, the performance of \mthree{} has very high accuracy (around 100\%) at the first layer for all models. Then, the accuracy is decreased to 72.07\%, 77.56\%, and 64.19\% for the top prediction of BERT, RoBERTa, and XLMR. The accuracy of the top two and two five predictions are above 81\% and 94\%. The Roberta model still has the best performance than the others, which has 77.56\%, 93.72\%, and 98.21\% accuracy for the top one, two, and five predictions (Table~\ref{tab:mnli_mapper_eval}). 

\onecolumn
\begin{table*}[!t]
\centering
\footnotesize
\begin{tabular}{l|c|c|c|c|c|c|c|c|c|c|c|c} 
\hline
         & \multicolumn{12}{c}{\textbf{MNLI}}                              \\ 
\hline
         & \multicolumn{4}{c|}{\textbf{BERT}}       &\multicolumn{4}{c|}{\textbf{RoBERTa}}    &\multicolumn{4}{c}{\textbf{XLM-R}}           \\ 
\hline
Layer    & \multicolumn{1}{l|}{0} & \multicolumn{1}{l|}{1} & \multicolumn{1}{l|}{2} &\multicolumn{1}{l|}{Mix} & \multicolumn{1}{l|}{0} & \multicolumn{1}{l|}{1} & \multicolumn{1}{l|}{2} &\multicolumn{1}{l|}{Mix} & \multicolumn{1}{l|}{0} & \multicolumn{1}{l|}{1} & \multicolumn{1}{l|}{2} &\multicolumn{1}{l}{Mix}   \\ 
\hline
Layer 0  & 0  & 6  & 0  & 394   &0 & 2  & 0 &398   & 0 & 7 & 0  & 393                      \\
Layer 1  & 0  & 4  & 0  & 396   &0 & 2  & 0 &398   & 0 & 4 & 0  & 396                      \\
Layer 2  & 0  & 3  & 0  & 397   &0 & 1  & 0 &399   & 0 & 3 & 0 & 397                      \\
Layer 3  & 0  & 4  & 0  & 396   &0 & 2  & 0 &398   & 0 & 5 & 0 & 395                      \\
Layer 4  & 0  & 4  & 0  & 396   &0 & 1  & 0 &399   & 0 & 4 & 0 & 396                      \\
Layer 5  & 0  & 4  & 0  & 396   &0 & 0  & 0 &400   & 0 & 4 & 0 & 396                      \\
Layer 6  & 0  & 6  & 0  & 394   &0 & 1  & 0 &399   & 0 & 4 & 0 & 396                      \\
Layer 7  & 0  & 4  & 0  & 396   &0 & 3  & 0 &397   & 0 & 2 & 0 & 398                      \\
Layer 8  & 0  & 1  & 0  & 399   &1 & 11 & 6 &382   & 0 & 1 & 0 & 399                      \\
Layer 9  & 0  & 1  & 0  & 399   &27 & 38 &24 &311  & 4 & 6 & 6 & 384                      \\
Layer 10 & 0  & 0  & 0  & 400   &38 &48 &34 &280   & 24 & 41 & 18 & 317                    \\
Layer 11 & 0  & 1  & 0  & 399   &51 &76 &50 &223   & 40 & 67 & 51 & 242                     \\
Layer 12 & 0  & 0  & 0  & 400   &92  &155 &81 &72  & 64 & 86 & 82 & 168                    \\
\bottomrule
\end{tabular}
\caption{Number of clusters for each polarity: '0' for entailment label, '1' for neutral label, and '2' for contradiction label. The total number of clusters is 400.}
\label{app:tab:cluster_tag_numbers_mnli}
\end{table*}

\begin{table*}[]
\footnotesize
\centering
\begin{tabular}{l|l|l|l|l|l|l|l|l|l}
    \toprule
        & \multicolumn{9}{c}{\textbf{MNLI}}  \\
        \midrule
        & \multicolumn{3}{c}{\textbf{BERT}} & \multicolumn{3}{c}{\textbf{RoBERTa}} & \multicolumn{3}{c}{\textbf{XLM-R}} \\
    \midrule
    Layer & Top-1 & Top-2 & Top-5 & Top-1 & Top-2 & Top-5 & Top-1 & Top-2 & Top-5 \\
    \midrule
0 & 100 & 100 & 100 & 99.97 & 100 & 100 & 100 & 100 & 100 \\
1 & 100 & 100 & 100 & 99.91 & 99.99 & 100 & 100 & 100 & 100 \\
2 & 100 & 100 & 100 & 99.92 & 99.99 & 100 & 99.75 & 100 & 100 \\
3 & 99.25 & 100 & 100 & 99.70 & 99.92 & 99.96 & 99.46 & 99.95 & 100\\
4 & 99.22 & 99.97 & 99.98 & 99.15 & 99.65 & 99.88 & 98.81 & 99.83 & 100 \\
5 & 99.04 & 99.95 & 99.99 & 97.07 & 96.98 & 99.26 & 97.72 & 99.42 & 99.89 \\
6 & 97.07 & 99.45 & 99.90 & 91.91 & 95.05 & 98.50 & 94.83 & 98.45 & 99.61 \\
7 & 96.81 & 99.35 & 99.85 & 96.99 & 87.21 & 95.70 & 86.96 & 95.37 & 98.72 \\
8 & 94.15 & 98.18 & 99.55 & 94.75 & 83.09 & 94.38 & 79.62 & 91.37 & 97.62 \\
9 & 90.08 & 96.52 & 98.90 & 91.52 & 82.80 & 94.38 & 73.73 & 88.57 & 96.76 \\
10 & 81.31 & 90.97 & 97.20 & 84.79 & 81.37 & 93.07 & 66.10 & 82.36 & 93.39 \\
11 & 79.05 & 89.62 & 96.51 & 81.79 & 91.09 & 98.10 & 68.30 & 84.49 & 95.28 \\
12 & 72.07 & 89.27 & 99.45 & 77.56 & 93.72 & 98.21 & 64.19 & 81.96 & 94.26 \\  \bottomrule
\end{tabular}
\caption{Top 1, 2, and 5 accuracy of ~\mthree{} in mapping a representation to the correct latent concept for the MNLI task. The top-5 performance reaches above 90\% for all models demonstrating that the correct latent concept is among the top probable latent concepts of \mthree{}.}
\label{tab:mnli_mapper_eval}
\end{table*}

\twocolumn
\section{LLama2}
\label{app:llama2}
\subsection{Experimental Setup}
We also tried the Eraser Movie sentiment classification and Jigsaw Toxicity classification tasks with the Llama2 model. We applied the ``Llama-2-7b-chat-hf'' version of the Llama2 model. We used the last token of the input prompt as the \texttt{[CLS]} token. We only used these \texttt{[CLS]} tokens as the latent concept explanation. For~\mone{}, we set $K=400$ for the sentiment and set $K=200$ for the toxicity.

\subsection{Sentiment Classification Task}

 \subsubsection{\mone{}}
Compared to the BERT, RoBERTa, and XLMR models (Table~\ref{app:tab:cluster_tag_numbers_eraser}), the Llama2 model has fewer class-based clusters at the last layer(See Table~\ref{app:tab:cluster_tag_numbers_llama_sentiment}). There are around 67\% class-based clusters detected at the last layer for the Llama2 model.

\begin{table}[]
\centering
\footnotesize
\begin{tabular}{l|c|c|c} 
\hline
         & \multicolumn{3}{c}{\textbf{Sentiment}}                              \\ 
\hline
         & \multicolumn{3}{c}{\textbf{Llama-2-7b-chat-hf}} \\ 
\hline
Layer    & \multicolumn{1}{l|}{Negative} & \multicolumn{1}{l|}{Positive} & \multicolumn{1}{l}{Mix} \\ 
\hline
Layer 0  & 27 & 372  & 1 \\
Layer 4  & 18 & 12  & 370 \\
Layer 8  & 21 & 21  & 358 \\
Layer 12  & 73 & 47  & 279  \\
Layer 16  & 154 & 90  & 155 \\
Layer 20  & 163 & 102  & 134 \\
Layer 24  & 173 & 108  & 118 \\
Layer 28 & 159 & 106 & 134 \\
Layer 32 & 164  & 103 & 132 \\
\hline
\end{tabular}
\caption{Number of clusters for each polarity. The total number of clusters is 400.}
\label{app:tab:cluster_tag_numbers_llama_sentiment}
\end{table}

\subsubsection{\mtwo{}}
With the Llama2 model, the accuracy in mapping the salient word representation to the correct latent concept for the last layer is approximately 70\% (See Table~\ref{tab:eraser_predicted_attribution_position_llama}). Although this accuracy indicates that the Llama2 model performs well, it is notably lower than the accuracy achieved by the \mtwo{} model based on BERT, RoBERTa, and XLMR models, which has significantly high performance (Table~\ref{tab:eraser_predicted_attribution_saliency}).

\begin{table}[]
\centering
\footnotesize
\begin{tabular}{l|c}
    \toprule
    & \multicolumn{1}{c}{\textbf{Sentiment}}  \\
    \midrule
    Layer & Llama-2-7b-chat-hf \\
    \midrule
Layer 0 & 2.88	\\
Layer 4	& 0.93	\\
Layer 8 & 1.94 \\
Layer 12 & 22.11\\
Layer 16 & 64.18\\
Layer 20 & 70.63 \\
Layer 24 & 75.64\\
Layer 28 & 71.30\\
Layer 32 & 71.02\\
    \midrule
\end{tabular}
\caption{Saliency-based method: accuracy of \mtwo{} in mapping a representation to the correct latent concept in the sentiment classification task using Llama2 model.}
\label{tab:eraser_predicted_attribution_position_llama}
\end{table}

\subsubsection{\mthree{}}
We found that, like the performance of using the other three models, the performance of \mthree{} using the Llama2 model exhibits a high Top-1 accuracy (97.55\%) in the lower layers, and decreases to 66.47\% for the last layer(Table~\ref{tab:sentimemt_mapper_eval_llama}). Additionally, the top two and five predictions of the mapper achieve accuracies of 82.84\% and 94.88\%, respectively. The accuracy of \mthree{} using the Llama2 model is relatively lower compared to its accuracy using BERT, RoBERTa, and XLM-RoBERTa(Table~\ref{tab:eraser_mapper_eval}).

\begin{table}[]
\footnotesize
\centering
\begin{tabular}{l|l|l|l}
    \toprule
        & \multicolumn{3}{c}{\textbf{Sentiment}}  \\
        \midrule
        & \multicolumn{3}{c}{\textbf{Llama-2-7b-chat-hf}}\\
    \midrule
    Layer & Top-1 & Top-2 & Top-5 \\
    \midrule
0 & 97.55 & 97.55 & 97.55 \\
4 & 19.90 & 31.36 & 47.08 \\
8 & 49.46 & 68.06 & 86.37 \\
12 & 60.85 & 77.43 & 92.36 \\
16 & 61.86 & 80.97 & 95.03 \\
20 & 64.02 & 80.61 & 94.23 \\
24 & 63.95 & 82.26 & 94.23 \\
28 & 65.83 & 81.25 & 94.52 \\
32 & 66.47 & 82.84 & 94.88 \\ \bottomrule
\end{tabular}
\caption{Top 1, 2, and 5 accuracy of ~\mthree{} in mapping a representation to the correct latent concept for the sentiment classification task using the Llama2 model. }
\label{tab:sentimemt_mapper_eval_llama}
\end{table}

\subsection{Toxicity Classification Task}
\subsubsection{\mone{}}
We found that 83\% of the latent concepts of Llama2 are the class label-based at the last layer(Table~\ref{app:tab:cluster_tag_numbers_llama_toxicity}). The BERT, RoBERTa, and XLMR models have a relatively higher number of class label-based clusters(Table~\ref{app:tab:cluster_tag_numbers_toxicity}).

\begin{table}[!htbp]
\centering
\footnotesize
\begin{tabular}{l|c|c|c} 
\hline
         & \multicolumn{3}{c}{\textbf{Toxicity}}                              \\ 
\hline
         & \multicolumn{3}{c}{\textbf{Llama-2-7b-chat-hf}} \\ 
\hline
Layer    & \multicolumn{1}{l|}{Non-toxic} & \multicolumn{1}{l|}{toxic} & \multicolumn{1}{l}{Mix} \\ 
\hline
Layer 0  & 84 & 108  & 1 \\
Layer 4  & 35 & 13  & 150 \\
Layer 8  & 27 & 5  & 168 \\
Layer 12  & 43 & 22  & 135  \\
Layer 16  & 61 & 21  & 117 \\
Layer 20  & 62 & 25  & 113 \\
Layer 24  & 69 & 25  & 106 \\
Layer 28 & 67 & 26 & 107 \\
Layer 32 & 69  & 21 & 109 \\
\hline
\end{tabular}
\caption{Number of clusters for each polarity. The total number of clusters is 200.}
\label{app:tab:cluster_tag_numbers_llama_toxicity}
\end{table}

\subsubsection{\mtwo{}}
The accuracy of the Llama2 model in our experiments is significantly lower compared to BERT, RoBERTa, and XLMR (Table~\ref{tab:toxicity_predicted_attribution_position_llama}). The performance of the other three models achieves accuracy values exceeding 90\% (Table~\ref{tab:toxicity_predicted_attribution_saliency}). The lower accuracy is due to several reasons. Llama2 is a generative model and it is hard to restrict its output to a single class. While we optimized the prompt for this purpose, we classified responses as label 0 (non-toxic) only if they contained ``non-toxic'', ``NON-TOXIC'', or ``Non-toxic''. Similarly, we classified responses as 1 (toxic) if they contained variations of the term ``toxic''. Moreover, many responses of the model did not provide a classification result due to inappropriate or disrespectful content of input instances that was blocked by the safety filter. Consequently, there are many sentences were skipped, which may account for the lower accuracy of Llama2 compared to the other models.

\begin{table}[]
\centering
\footnotesize
\begin{tabular}{l|c}
    \toprule
    & \multicolumn{1}{c}{\textbf{Toxicity}}  \\
    \midrule
    Layer & Llama-2-7b-chat-hf \\
    \midrule
Layer 0 & 2.26	\\
Layer 4	& 7.20	\\
Layer 8 & 6.59 \\
Layer 12 & 32.10 \\
Layer 16 & 42.91\\
Layer 20 & 45.83 \\
Layer 24 & 46.93\\
Layer 28 & 46.43\\
Layer 32 & 44.28\\
    \midrule
\end{tabular}
\caption{Saliency-based method: accuracy of \mtwo{} in mapping a representation to the correct latent concept in the toxicity classification task using Llama2 model.}
\label{tab:toxicity_predicted_attribution_position_llama}
\end{table}

\subsubsection{\mthree{}}
The top-1 performance of~\mthree{} based on the Llama2 model achieves 74.44\% for the last layer(Table~\ref{tab:toxicity_mapper_eval_llama}). This performance is better than the one based on the BERT and XLM-Roberta (Table~\ref{tab:toxicity_mapper_eval}). RoBERTa still delivers the best performance.

\begin{table}[]
\footnotesize
\centering
\begin{tabular}{l|l|l|l}
    \toprule
        & \multicolumn{3}{c}{\textbf{Toxicity}}  \\
        \midrule
        & \multicolumn{3}{c}{\textbf{Llama-2-7b-chat-hf}}\\
    \midrule
    Layer & Top-1 & Top-2 & Top-5 \\
    \midrule
0 & 96.97 & 96.97 & 97.09 \\
4 & 42.38 & 62.00 & 83.86 \\
8 & 67.83 & 85.20 & 97.20 \\
12 & 70.40 & 89.24 & 98.21 \\
16 & 73.09 & 87.44 & 98.77 \\
20 & 74.22 & 90.25 & 98.99 \\
24 & 71.19 & 88.68 & 98.88 \\
28 & 72.65 & 90.13 & 98.76 \\
32 & 74.44 & 91.82 & 99.10 \\ \bottomrule
\end{tabular}
\caption{Top 1, 2, and 5 accuracy of ~\mthree{} in mapping a representation to the correct latent concept for the toxicity classification task using the Llama2 model. }
\label{tab:toxicity_mapper_eval_llama}
\end{table}

\end{document}